\documentclass[journal]{IEEEtai}

\usepackage[colorlinks,urlcolor=blue,linkcolor=blue,citecolor=blue]{hyperref}

\usepackage{color,array}

\usepackage{graphicx}

\usepackage{subcaption}
\usepackage{cite}

\usepackage{multirow}

\begin{document}

\title{Vision-Based Learning for Drones: A Survey}

\author{Jiaping Xiao, 
%\IEEEmembership{Graduate Student Member, IEEE}%,
Rangya Zhang, Yuhang Zhang, and Mir Feroskhan, \IEEEmembership{Member, IEEE}
% \thanks{This paragraph of the first footnote will contain the date on which you submitted your paper for review. It will also contain support information, including sponsor and financial support acknowledgment. For example, ``This work was supported in part by the U.S. Department of Commerce under Grant BS123456.''}
\thanks{J. Xiao, R. Zhang, Y. Zhang and M. Feroskhan are with the School of Mechanical and Aerospace Engineering, Nanyang Technological University, Singapore 639798, Singapore (e-mail: jiaping001@e.ntu.edu.sg; rangya001@e.ntu.edu.sg; yuhang002@e.ntu.edu.sg; mir.feroskhan@ntu.edu.sg). \textit{(Corresponding author: Mir Feroskhan.)}}
% \thanks{S. B. Author, Jr., was with Rice University, Houston, TX 77005 USA. He is now with the Department of Physics, Colorado State University, Fort Collins, CO 80523 USA (e-mail: author@lamar.colostate.edu).}
% \thanks{T. C. Author is with the Electrical Engineering Department, University of Colorado, Boulder, CO 80309 USA, on leave from the National Research Institute for Metals, Tsukuba, Japan (e-mail: author@nrim.go.jp).}
% \thanks{This paragraph will include the Associate Editor who handled your paper.}
}

\markboth{}
{}
% \markboth{Journal of IEEE Transactions on Artificial Intelligence, Vol. 00, No. 0, Month 2020}
% {First A. Author \MakeLowercase{\textit{et al.}}: Bare Demo of IEEEtai.cls for IEEE Journals of IEEE Transactions on Artificial Intelligence}

\maketitle

\begin{abstract}
Drones as advanced cyber-physical systems are undergoing a transformative shift with the advent of vision-based learning, a field that is rapidly gaining prominence due to its profound impact on drone autonomy and functionality. Different from existing task-specific surveys, this review offers a comprehensive overview of vision-based learning in drones, emphasizing its pivotal role in enhancing their operational capabilities under various scenarios. We start by elucidating the fundamental principles of vision-based learning, highlighting how it significantly improves drones' visual perception and decision-making processes. We then categorize vision-based control methods into indirect, semi-direct, and end-to-end approaches from the perception-control perspective. We further explore various applications of vision-based drones with learning capabilities, ranging from single-agent systems to more complex multi-agent and heterogeneous system scenarios, and underscore the challenges and innovations characterizing each area. Finally, we explore open questions and potential solutions, paving the way for ongoing research and development in this dynamic and rapidly evolving field. With growing large language models (LLMs) and embodied intelligence, vision-based learning for drones provides a promising but challenging road towards artificial general intelligence (AGI) in 3D physical world.

% This review aims to provide a thorough understanding of the current state and future prospects of vision-based learning in drones.
\end{abstract}

\begin{IEEEkeywords}
% Enter key words or phrases in alphabetical order, separated by commas. For a list of suggested keywords, send a blank e-mail to \href{mailto:keywords@ieee.org}{\underline{keywords@ieee.org}} or visit \href{http://www.ieee.org/organizations/pubs/ani_prod/keywrd98.txt}{\underline{http://www.ieee.org/organizations/pubs/ani\_prod/keywrd98.txt}}
Drones, learning systems, robotics learning, embodied intelligence.
\end{IEEEkeywords}

\section{Introduction}

\IEEEPARstart{D}{rones} are intelligent cyber-physical systems (CPS) \cite{rajkumar2010cyber,baheti2011cyber} that rely on functional sensing, system communication, real-time computation, and flight control to achieve perception and autonomous flight. Due to their high autonomy and maneuverability, drones \cite{hassanalian2017classifications} have been widely used in various missions, such as industrial inspection \cite{nooralishahi2021drone}, precision agriculture \cite{stehr2015drones}, parcel delivery \cite{9091085,kornatowski2020morphing} and search-and-rescue \cite{camara2014cavalry} (some applications are shown in Fig. \ref{fig:application}). As such, drones in smart cities are an essential feature of the service industry of tomorrow. For future applications, many novel drones and functions are under development with the advent of advanced materials (adhesive and flexible materials), miniaturization of electronic and optical components (sensors, microprocessors), onboard computers (Nividia Jetson, Intel NUC, Raspeberry Pi, etc.), batteries, and localization systems (SLAM, UWB, GPS, etc.). Meanwhile, the functionality of drones is becoming more complex and intelligent due to the rapid advancement of artificial intelligence (AI) and onboard computation capability. From the perspective of innovation, there are three main directions for the future development of drones. Specifically:

\begin{figure}[t]
    \centering
    \includegraphics[width=3.4in]{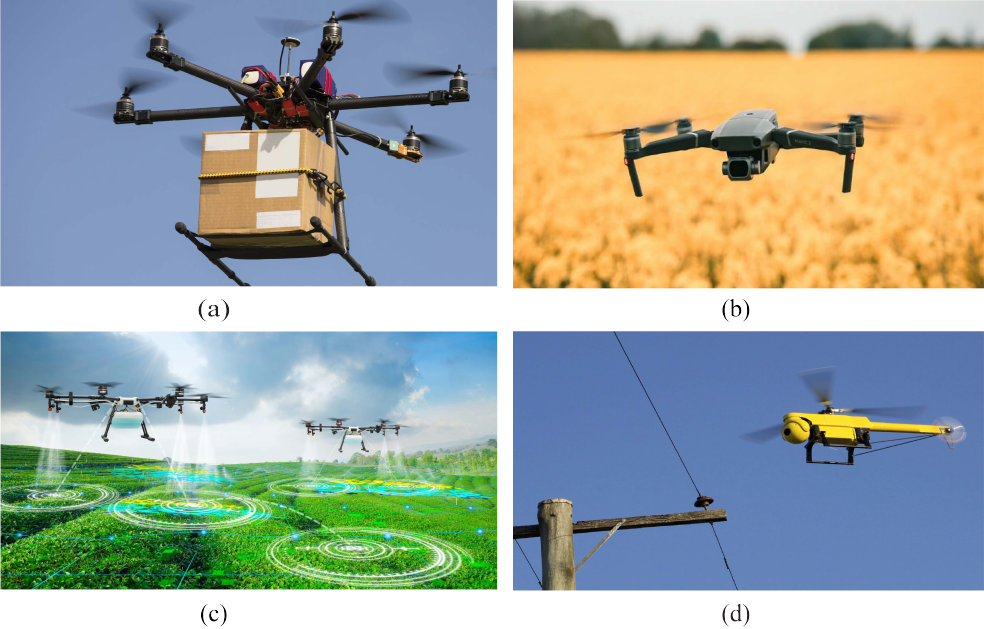}
    \caption{Applications of vision-based drones. (a) Parcel delivery; (b) Photography; (c) Precision agriculture; (d) Power grid inspection.}
    \label{fig:application}
\end{figure}

\begin{itemize}
\item{\textbf{The minimization of drones}.} Micro and nano drones are capable of accomplishing missions at a low cost and without space constraints. The small autonomous drones raised the interest of scientists to obtain more inspiration from biology, such as bees \cite{graule2016perching} and flipping birds \cite{floreano2015science}.

\item{\textbf{The novel designs of drones}.} Structure and aerodynamic design enable drones to obtain increased maneuverability and improved flight performance. Tilting, morphing, and folding structures and actuators are widely studied in drone design and control \cite{shu2019quadrotor, kornatowski2020morphing, ajanic2020bioinspired}. 

\item{\textbf{The autonomy of drones}.} Drone autonomy achieves autonomous navigation and task execution. It requires real-time perception and onboard computation capabilities. Drones that integrate visual sensors, efficient online planning, and learning-based decision-making algorithms have notably enhanced autonomy and intelligence \cite{zhou2020ego, kaufmann2020deep}, even beating world-level champions in drone racing with a vision-based learning system called Swift \cite{kaufmann2023champion}.
\end{itemize}

\begin{figure}[t]
    \centering
    \includegraphics[width=3.4in]{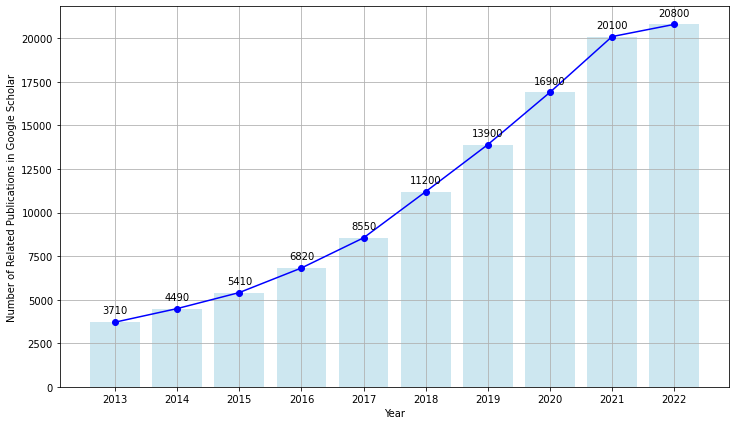}
    \caption{Number of related publications in Google Scholar using keyword ``vision-based learning drones".}
    \label{fig:literature}
\end{figure}

Recently, to improve drone autonomy, vision-based learning drones, which combine advanced sensing with learning capabilities, are attracting more insights (see the rapid growth trend in Fig. \ref{fig:literature}). With such capabilities, drones are even emerging towards artificial general intelligence (AGI) in the 3D physical world, especially when integrated with rapid-growing large language models (LLMs) \cite{singh2023progprompt} and embodied intelligence \cite{gupta2021embodied}. Existing vision-based drone-related surveys focus only on specific tasks and applications, such as UAV navigation \cite{lu2018survey, arafat2023vision}, vision-based UAV landing \cite{kakaletsis2021computer}, obstacle avoidance \cite{mcfadyen2016survey}, vision-based inspection with UAVs \cite{jenssen2018automatic, spencer2019advances}, and autonomous drone racing \cite{hanover2023autonomous}, which limits the understanding of vision-based learning in drones from a holistic perspective. Therefore, this survey provides a comprehensive review of vision-based learning for drones to deliver a more general view of current drone autonomy technologies, including background, visual perception, vision-based control, applications and challenges, and open questions with potential solutions. To summarize, our main contributions are:
\begin{itemize}
    \item We discussed the development of vision-based drones with learning capabilities and analyzed the core components, especially visual perception and machine learning applied in drones. We further highlighted object detection with visual perception and how it benefits drone applications.
    \item We discussed the current state of vision-based control methods for drones and categorized them into indirect, semi-direct, and end-to-end methods from the perception-control perspective. This perspective helps to understand vision-based control methods and differentiate them with better features.
    \item We summarized the applications of vision-based learning drones in single-agent systems, multi-agent systems, and heterogeneous systems and discussed the corresponding challenges in different applications.
    \item We explored several open questions that can hinder the development and applicability of vision-based learning for drones. Furthermore, we discuss potential solutions for each question.
\end{itemize}

\textit{Organization:} The rest of this survey is organized as follows: Section \ref{sec:back} discusses the concept of vision-based learning drones and their core components; Section \ref{sec:obj} summarizes object detection with visual perception and its application to vision-based drones; We introduce the vision-based control methods for drones and categorize them in Section \ref{sec:vision-control}; The applications and challenges of vision-based learning for drones are discussed in Section \ref{sec:app}; We list the open questions faced by vision-based learning drones and potential solutions in Section \ref{sec: question}; Section \ref{sec:conclusion} summarizes and concludes this survey.

\begin{figure}[t]
    \centering
    \includegraphics[width=3.4in]{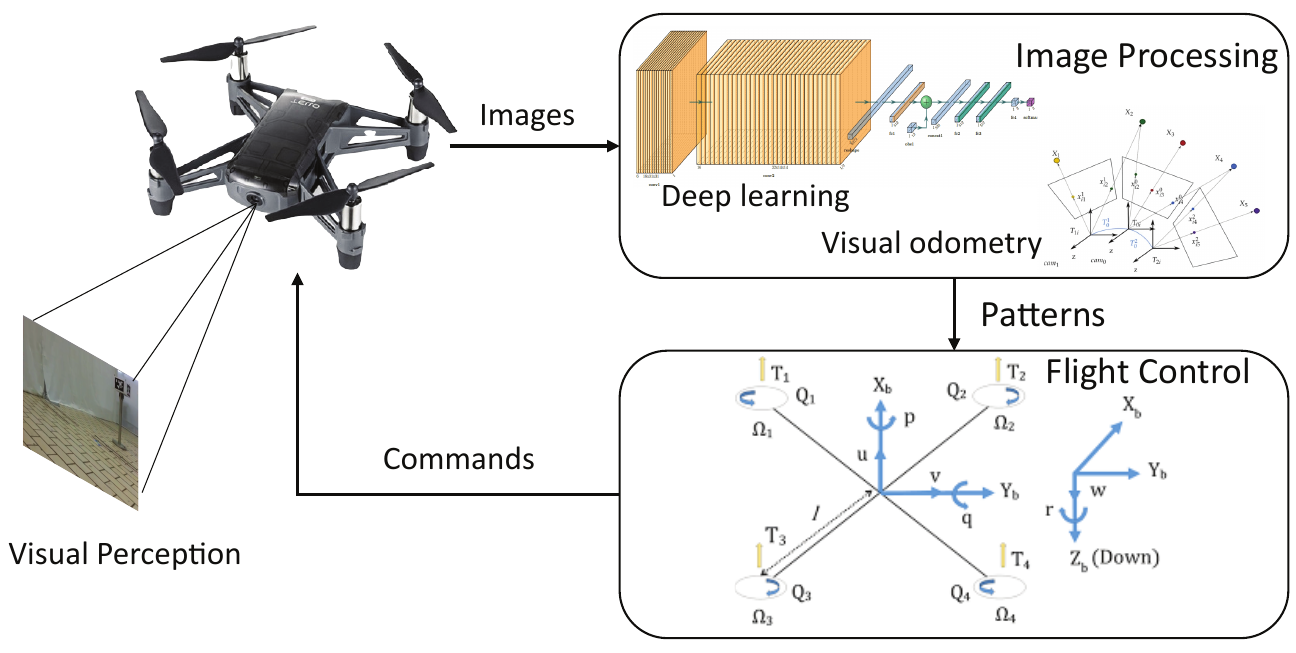}
    \caption{General framework of vision-based drones}
    \label{fig:vision-drone}
\end{figure}

\section{Background} \label{sec:back}
\subsection{Vision-based Learning Drones}
\begin{figure}[!b]
\centering
\begin{subfigure}{0.23\textwidth}
    \centering
  % include first image
    \centering
    \includegraphics[height=0.58\linewidth]{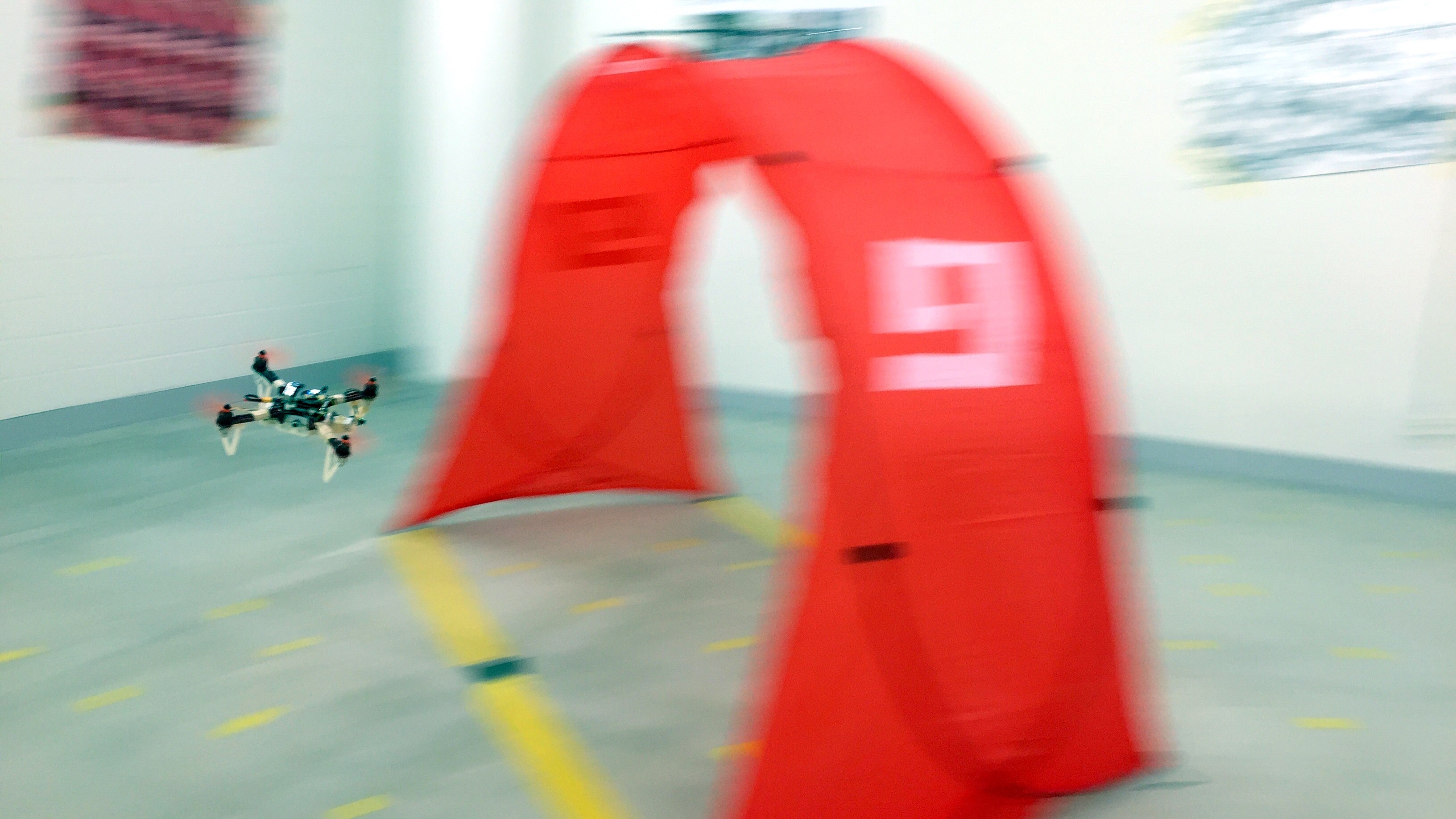}
    \caption{}
\end{subfigure}
\begin{subfigure}{0.23\textwidth}
  \centering
  % include second image
  \includegraphics[height=0.58\linewidth]{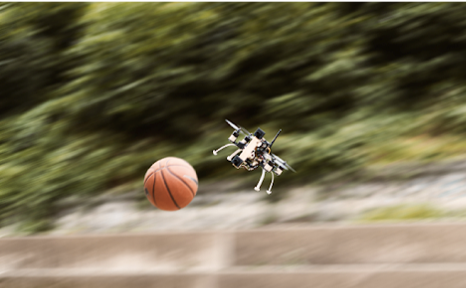}
  \caption{}
\end{subfigure}
\caption{Vision-based control for drones' obstacle avoidance in simple dynamic environments. (a) Drone racing in a dynamic environment with moving gates  \cite{kaufmann2018deep}; (b) A drone avoiding a ball thrown to it with event cameras \cite{falanga2020dynamic}.}
\label{dynamic_env}
\end{figure}

A typical vision-based drone consists of three parts (see Fig. \ref{fig:vision-drone}): (1) \textbf{Visual perception}: sensing the environment around the drone via monocular cameras or stereo cameras; (2) \textbf{Image processing}: extracting features from an observed image sequence and output specific patterns or information, such as navigation information, depth information, object information; (3) \textbf{Flight controller}: generating high-level and low-level commands for drones to perform assigned missions. Image processing and flight controllers are generally conducted on the onboard computer, while visual perception relies on the performance of visual sensors. Vision-based drones have been widely used in traditional missions such as environmental exploration \cite{10038280}, navigation \cite{zhou2020ego}, and obstacle avoidance \cite{kaufmann2018deep, falanga2020dynamic}. With efficient image processing and simple path planners, they can avoid dynamic obstacles effectively (see Fig. \ref{dynamic_env}).

Currently, vision-based learning drones, which utilize visual sensors and efficient learning algorithms, have achieved remarkable advanced performance in a series of standardized visual perception and decision-making tasks, such as agile flight control \cite{Loquercio2021}, navigation \cite{RN10} and obstacle avoidance \cite{falanga2020dynamic}. Various cases have showcased the power of learning algorithms in improving the agility and perception capabilities of vision-based drones. For instance, using only depth cameras, inertial measurement units (IMU) and a lightweight onboard computer, the vision-based learning drone in \cite{Loquercio2021} succeeded in performing high-speed flight in unseen and unstructured wild environments. The controller of the drone in \cite{Loquercio2021} was trained from a high-fidelity simulation and transferred to a physical platform. Following that, the Swift system was developed in \cite{kaufmann2023champion} to achieve world champion-level autonomous drone racing with a tracking camera and a shallow neural network. Such kinds of vision-based learning drones are leading the future of drones due to their perception and learning capabilities in complex environments. Similarly, with event cameras and a trained narrow neural network-EVDodgeNet, the vision-based learning drone in \cite{sanket2020evdodgenet} was able to dodge multiple dynamic obstacles (balls) during flight. Afterwards, to improve the perception onboard in the real world, an uncertainty estimation module was trained with the Ajna network \cite{sanket2023ajna}, which significantly increased the generalization capability of learning-based control for drones. The power of deep learning in handling uncertain information frees traditional approaches from complex computation with necessary accurate modeling.

\subsection{Visual Perception}
Visual perception for drones is the ability of drones to perceive their surroundings and their own states through the extraction of necessary features for specific tasks with visual sensors. Light detection and ranging (LIDAR) and cameras are commonly used sensors to perceive the surrounding environment for drones.

\begin{figure}[!tb]
\begin{subfigure}{0.24\textwidth}
    \centering
  % include first image
    \centering
    \includegraphics[width=.9\linewidth]{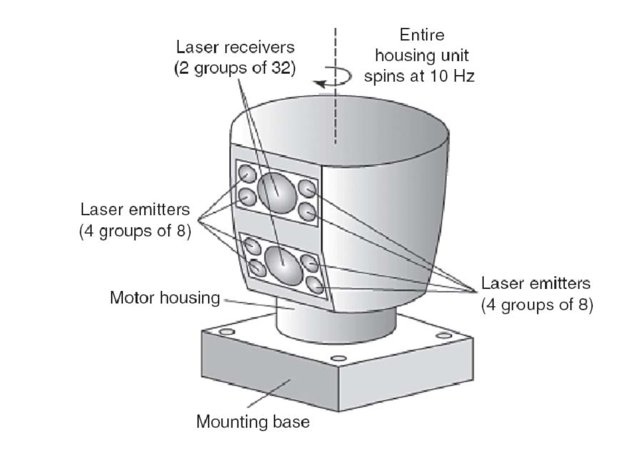}
    \caption{}
    \label{fig5}
\end{subfigure}
\begin{subfigure}{0.24\textwidth}
  \centering
  % include second image
  \includegraphics[width=.9\linewidth]{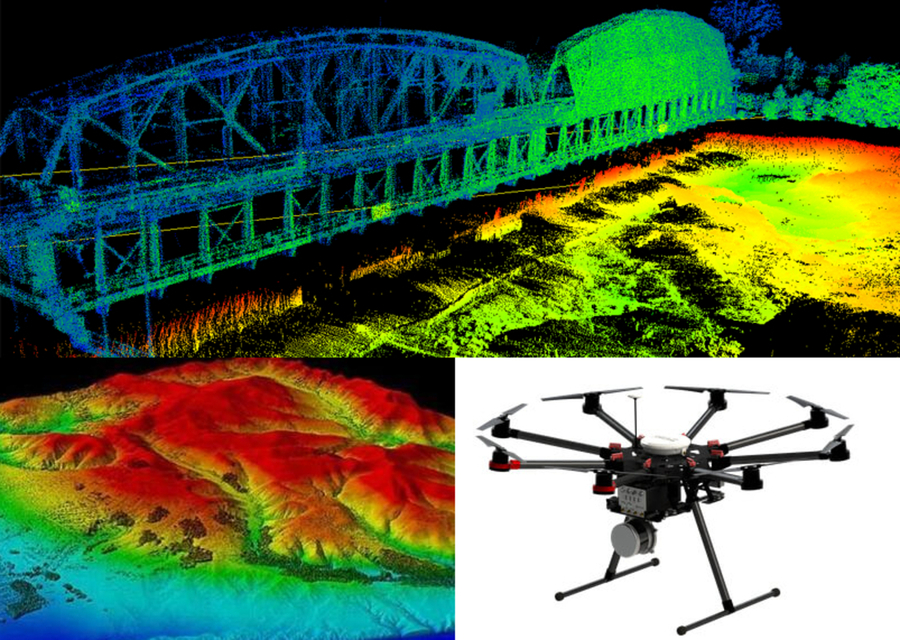}  
  \caption{}
  \label{fig:sub-second}
\end{subfigure}
\caption{LIDAR on drones for visual perception. (a) A typical surrounding LIDAR; (b) Generated point cloud with LIDAR.}
\label{lidar}
\end{figure}

\subsubsection{LIght Detection And Ranging (LIDAR)}
LIDAR is a kind of active range sensor that relies on the calculated time of flight (TOF) between the transmitted and received beams (laser) to estimate the distance between the robot and the reflected surface of objects \cite{siegwart2011introduction}. Based on the scanning mechanism, LIDAR can be divided into solid-state LIDAR, which has a fixed field of view without moving parts, and surrounding LIDAR, which spins to provide a 360-degree horizontal view. Surrounding LIDAR is also referred to as ``laser scanning" or ``3D scanning", which creates a 3D representation of the explored environment using eye-safe laser beams. A typical LIDAR (see Fig. \ref{lidar}) consists of laser emitters, laser receivers, and a spinning motor. The vertical field of view (FOV) of a LIDAR is determined by the number of vertical arrays of lasers. For instance, a vertical array of 16 lasers scanning 30 degrees gives a vertical resolution of 2 degrees in a typical configuration. LIDAR has recently been used on drones for mapping \cite{Michael2012}, power grid inspection \cite{Guan2021}, pose estimation \cite{opromolla2015uncooperative} and object detection \cite{wang2021multi}. LIDAR provides sufficient and accurate depth information for drones to navigate in cluttered environments. However, it is bulky and power-hungry and does not fit within the payload restrictions of agile autonomous drones. Meanwhile, using raycast representation in a simulation environment makes it hard to match the inputs of a real LIDAR device, which brings many challenges for the Sim2Real transfer when a learning approach is considered.

\begin{figure}[!tb]
\centering
\begin{subfigure}{0.35\textwidth}
    \centering
  % include first image
    \centering
    \includegraphics[width=.9\linewidth]{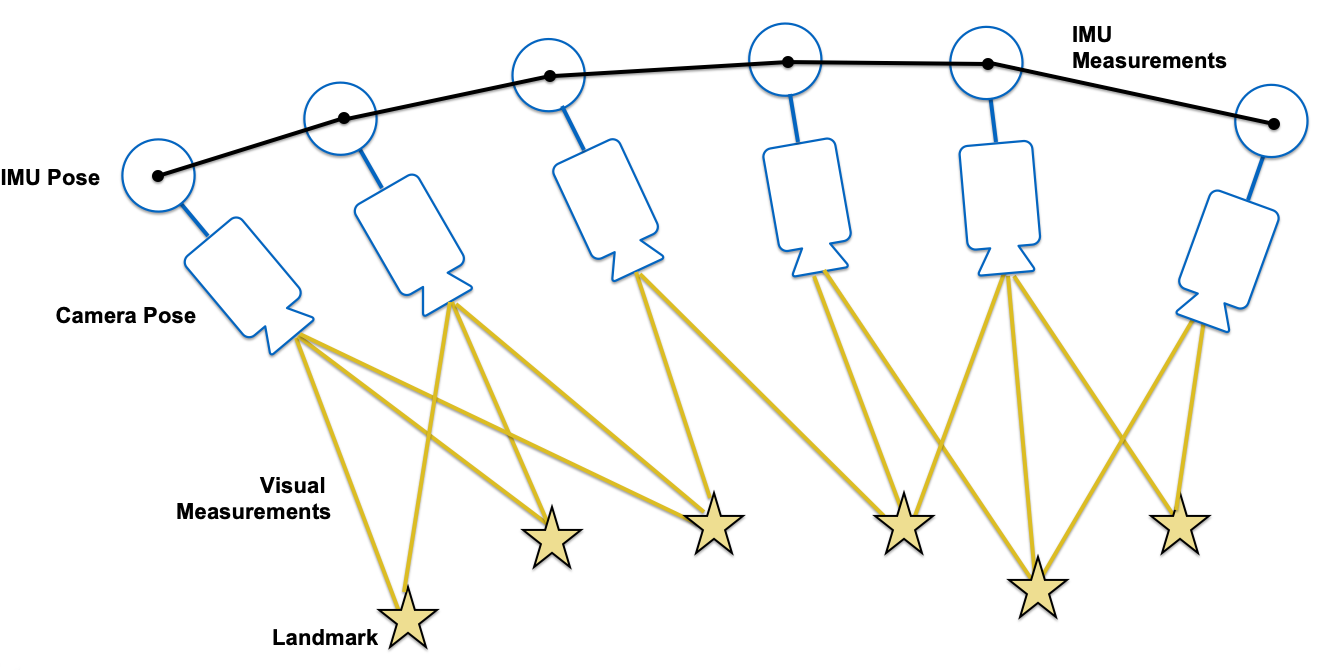}
    \caption{}
\end{subfigure}
\begin{subfigure}{.35\textwidth}
  \centering
  % include second image
  \includegraphics[width=.9\linewidth]{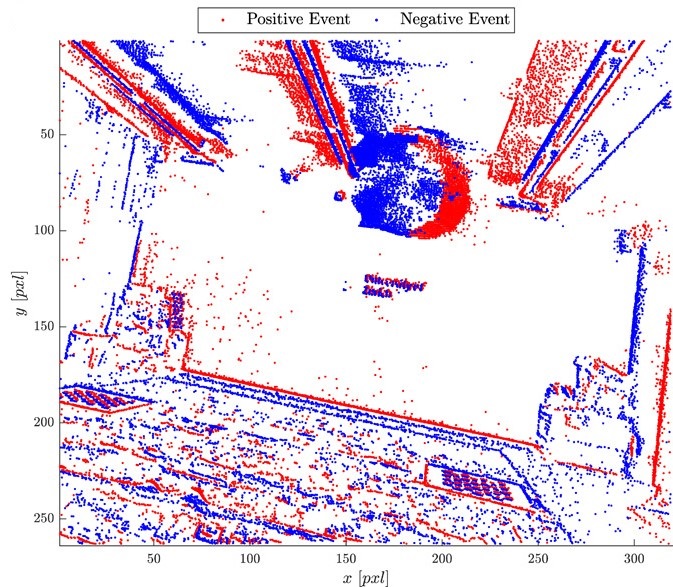}
  \caption{}

\end{subfigure}
\caption{Visual perception with cameras for drones. (a) Visual Odometry for drones' positioning; (b) Object detection for drones' obstacle avoidance with event camera \cite{falanga2020dynamic}. }
\label{camera}
\end{figure}

\subsubsection{Camera}
Compared to LIDAR, cameras provide a cheap and lightweight way for drones to perceive the environment. Cameras are external passive sensors used to monitor the drone's geometric and dynamic relationship to its task, environment or the objects that it is handling. Cameras are commonly used perception sensors for drones to sense environment information, such as objects' position and a point cloud map of the environment. In contrast to the motion capture system, which can only broadcast global geometric and dynamic pose information within a limited space from an offboard synchronized system, cameras enable a drone to fly without space constraints. Cameras can provide positioning for drone navigation in GPS-denied environments via visual odometry (VO) \cite{Aqel2016,Delmerico2018,Scaramuzza2020} and visual simultaneous localization and mapping systems (V-SLAM) \cite{mur2015orb,Qin2018, Campos2021}. Meanwhile, object detection and depth estimation can be performed with cameras to obtain the relative positions and sizes of obstacles. However, to avoid dynamic obstacles, even physical attacks like bird chasing, the agile vision-based drone poses fundamental challenges to visual perception. Motion blur, sparse texture environments, and unbalanced lighting conditions can cause the loss of feature detection in the VIO and object detection. LIDAR and event cameras \cite{gallego2020event} can partially address these challenges. However, LIDAR and event cameras are either too bulky or too expensive for agile drone applications. Considering the agility requirement of physical attack avoidance, lightweight dual-fisheye cameras are used for visual perception. With dual fisheye cameras, the drone can achieve better navigation capability \cite{gao2020autonomous} and omnidirectional visual perception \cite{kumar2021omnidet}. Some sensor fusion and state estimation techniques are required to alleviate the accuracy loss brought by the motion blur.

\subsection{Machine Learning}
Recently, machine learning (ML), especially deep learning (DL), has attracted much attention from various fields and has been widely applied to robotics for environmental exploration \cite{Haumann2010,Tan2022}, navigation in unknown environments \cite{zhu2017target,Mirowski2017,Wu2021}, obstacle avoidance, and intelligent control \cite{openai2019rubiks}. In the domain of drones, learning-based methods have also achieved promising success, particularly for deep reinforcement learning (DRL) \cite{pham2018autonomous, wu2019uav,xiao2021flying, Song2021,Loquercio2021, kaufmann2023champion, xiao2022collaborative}. In \cite{xiao2021flying}, a curriculum learning augmented end-to-end reinforcement learning was proposed for a UAV to fly through a narrow gap in the real world. A vision-based end-to-end learning method was successfully developed in \cite{Loquercio2021} to fly agile quadrotors through complex wild and human-made environments with only onboard sensing and computation capabilities, such as depth information. A visual drone swarm was developed in \cite{xiao2022collaborative} to perform collaborative target search with adaptive curriculum embedded multistage learning. These works verified the marvelous power of learning-based methods on drone applications, which pushes the agility and cooperation of drones to a level that classical approaches can hardly reach. Different from the classical approaches relying on separate mapping, localization, and planning, learning-based methods map the observations, such as the visual information or localization of obstacles, to commands directly without further planning. This greatly helps drones handle uncertain information in operations. However, learning-based methods require massive experiences and training datasets to obtain good generalization capability, which poses another challenge in deployment over unknown environments.

\section{Object Detection with Visual Perception} \label{sec:obj}
Object detection is a pivotal module in vision-based learning drones when handling complex missions such as inspection, avoidance, and search and rescue. Object detection is to find out all the objects of interest in the image and determine their position and size \cite{8627998}. Object detection is one of the core problems in the field of computer vision (CV). Nowadays, the applications of object detection include face detection, pedestrian detection, vehicle detection, and terrain detection in remote sensing images. Object detection has always been one of the most challenging problems in the field of CV due to the different appearances, shapes, and poses of various objects, as well as the interference of factors such as illumination and occlusion during imaging. At present, the object detection algorithm can be roughly divided into two categories: \textbf{multi-stage (two-stage) algorithm}, whose idea is to first generate candidate regions and then perform classification, and \textbf{one-stage algorithm}, the idea of which is to directly apply the algorithm to the input image and output the categories and corresponding positions. Beyond that, to retrieve 3D positions, depth estimation has been a popular research subbranch related to object detection whether using monocular \cite{bhat2023zoedepth} or stereo depth estimation \cite{laga2020survey}. For a very long time, the core neural network module (backbone) of object detection has been the convolutional neural network (CNN) \cite{9451544}. CNN is a classic neural network in image processing that originates from the study of the human optic nerve system. The main idea is to convolve the image with the convolution kernel to obtain a series of reorganization features, and these reorganization features represent the important information of the image. As such, CNN not only has the ability to recognize the image but also effectively decreases the requirement for computing resources. Recently, vision transformers (ViTs) \cite{dosovitskiy2020image}, originally proposed for image classification tasks, have been extended to the realm of object detection \cite{liu2021swin}. These models demonstrate superior performance by utilizing the self-attention mechanism, which processes visual information non-locally \cite{10088164}. However, a major limitation of ViTs is their high computational demand. This presents difficulties in achieving real-time inference, particularly on platforms with limited resources like drones.

\begin{figure}[!tbp]
\centering
\begin{subfigure}{0.5\textwidth}
    \centering
  % include first image
    \centering
    \includegraphics[width=.9\linewidth]{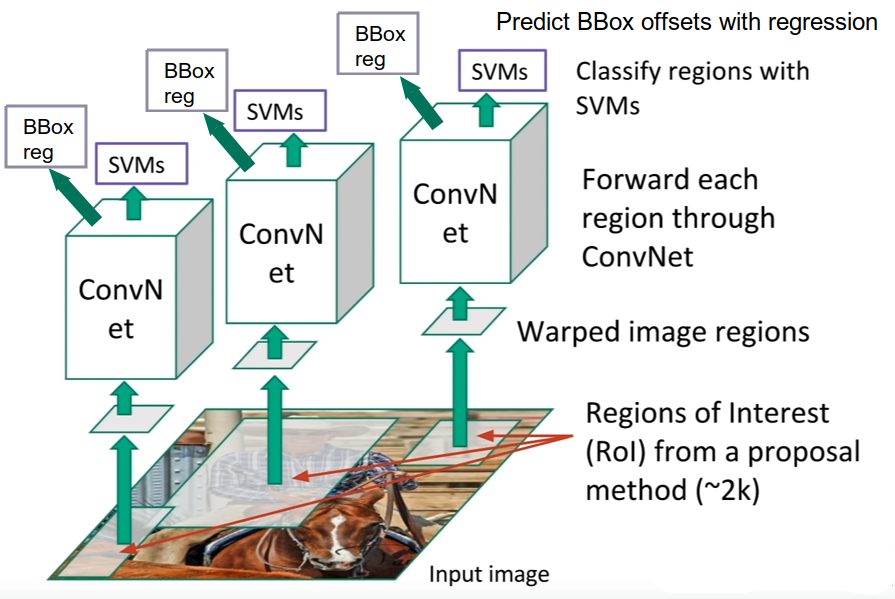}
    \caption{}
    \label{rcnn}
\end{subfigure}
\begin{subfigure}{0.5\textwidth}
  \centering
  % include second image
  \includegraphics[width=.9\linewidth]{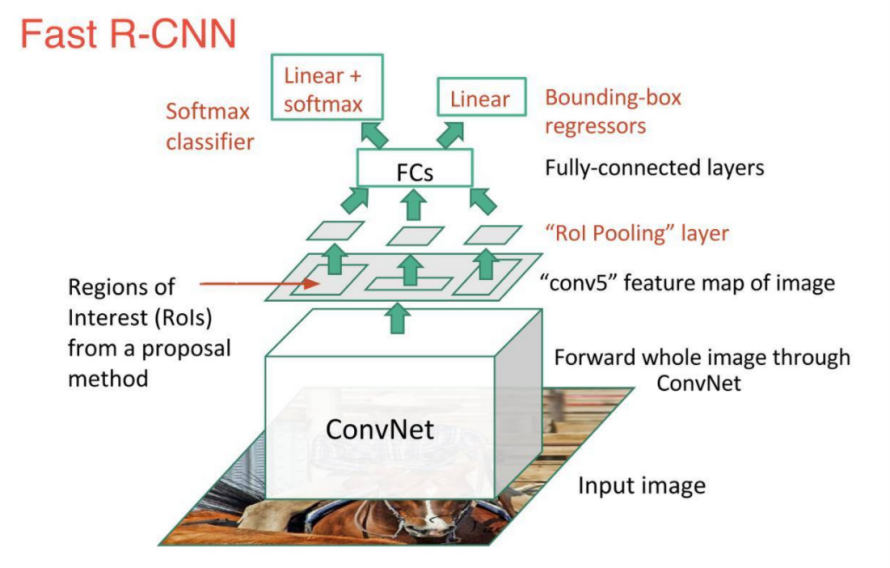}  
  \caption{}
  \label{fast-rcnn}
\end{subfigure} 

\caption{(a) R-CNN neural network architecture \cite{girshick2015region}; (b) Fast R-CNN neural network architecture \cite{girshick2015fast}.}
\label{rcnn2}
\end{figure}

\subsection{Multi-stage Algorithms} 
Classic multi-stage algorithms include RCNN (Region-based Convolutional Neural Network) \cite{girshick2015region}, Fast R-CNN \cite{girshick2015fast}, Faster R-CNN \cite{ren2015faster}. Multi-stage algorithms can basically meet the accuracy requirements in real-life scenarios, but the model is more complex and cannot be really applied to scenarios with high-efficiency requirements. In the R-CNN structure \cite{girshick2015region}, it is necessary to first give some regional proposals (RP), then use the convolutional layer for feature extraction, and then classify the regions according to these features. That is, the object detection problem is transformed into an image classification problem. The R-CNN model is very intuitive, but the disadvantage is that it is too slow, and the output is obtained via training multiple Support Vector Machines (SVMs). To solve the problem of slow training speed, the Fast R-CNN model is proposed (Fig. \ref{fast-rcnn}). This model has two improvements to R-CNN: (1) first use the convolutional layer to perform feature selection on the image so that only one convolutional layer can be used to obtain RP; (2) convert training multiple SVMs to use only one fully-connected layer and a softmax layer. These techniques greatly improve the computation speed but still fail to address the efficiency issue of the Selective Search Algorithm (SSA) for RP.

Faster R-CNN is an improvement on the basis of Fast R-CNN. In order to solve the problem of SSA, the SSA that generates RP in Fast R-CNN is replaced by a Region Proposal Network (RPN) and uses a model that integrates RP generation, feature extraction, object classification and object box regression. RPN is a fully convolutional network that simultaneously predicts object boundaries at each location. RPN is trained end-to-end to generate high-quality region proposals, which are then detected by Fast R-CNN. At the same time, RPN and Fast R-CNN share convolutional features. Meanwhile, in the feature extraction stage, Faster R-CNN uses a convolutional neural network. The model achieves $73.2\%$ and $70.4\%$ mean Average Precision (mAP) per category on the PASCAL VOC 2007 and 2012 datasets, respectively. Faster R-CNN has been greatly improved in speed than Fast R-CNN, and the accuracy has reached the state-of-the-art (SOTA), and it also fully developed an end-to-end object detection framework. However, Faster R-CNN still cannot achieve real-time object detection. Besides, after obtaining RP, it requires heavy computation for each RP classification.

\begin{figure}[htbp]

  \centering
  % include second image
  \includegraphics[width=3.0in]{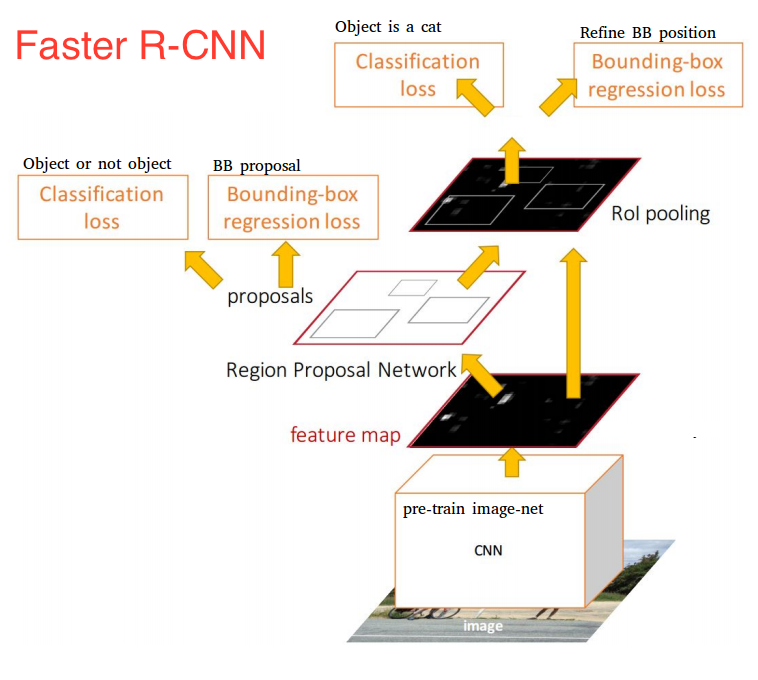}  
  \caption{Faster RCNN neural network architecture \cite{ren2015faster}.}
  \label{faster-rcnn}

\end{figure}

\subsection{One-stage Algorithms} 
\begin{figure}[htbp]

  \centering
  % include second image
  \includegraphics[width=1.0\linewidth]{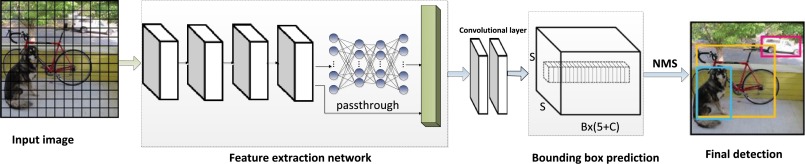}  
  \caption{YOLO network architecture \cite{redmon2016you}.}
  \label{yolo}

\end{figure}

One-stage algorithms such as the Single Shot Multibox Detector (SSD) model \cite{liu2016ssd} and the YOLO series models \cite{redmon2016you} are generally slightly less accurate than the two-stage algorithms, but have simpler architectures, which can facilitate end-to-end training and are more suitable for real-time object detection. The basic process of YOLO (see Fig. \ref{yolo}) is divided into three phases, namely, zooming the image, passing the image through a full convolutional neural network, and using maximum value suppression (NMS). The main advantages of the YOLO model are that it is fast, with few background errors via global processing, and it has good generalization performance. Meanwhile, YOLO can formulate the detection task as a unified, end-to-end regression problem, and simultaneously obtain the location and classification by processing the image only once. But there are also some problems with the YOLO, such as rough mesh, which will limit YOLO's performance over small objects. However, the subsequent YOLOv3, YOLOv5, YOLOX \cite{ge2021yolox} and YOLOv8 improved the network on the basis of the original YOLO and achieved better detection results.

\begin{figure}[htbp]

  \centering
  % include second image
  \includegraphics[width=1.0\linewidth]{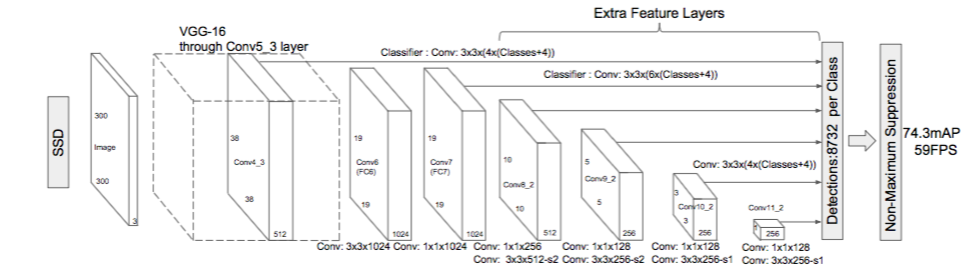}  
  \caption{SSD network architecture \cite{liu2016ssd}.}
  \label{ssd}

\end{figure}

SSD is another classic one-stage object detection algorithm. The flowchart of SSD is (1) first to extract features from the image through a CNN, (2) generate feature maps, (3) extract feature maps of multiple layers, and then (4) generate default boxes at each point of the feature map. Finally, (5) all the generated default boxes are collected and filtered using NMS. The neural network architecture of SSD is shown in Fig. \ref{ssd}. SSD uses more convolutional layers than YOLO, which means that SSD has more feature maps. At the same time, SSD uses different convolution segments based on the VGG model to output feature maps to the regressor, which tries to improve the detection accuracy over small objects.

The aforementioned multi-stage algorithms and one-stage algorithms have their own advantages and disadvantages. Multi-stage algorithms achieve high detection accuracy, but bring more computing overhead and repeated detection. The one-stage model generally consists of a basic network (Backbone Network) and a Detection Head. The former is used as a feature extractor to give representations of different sizes and abstraction levels of images; the latter one learns classification and location associations based on these representations and a supervised dataset. The two tasks of category prediction and position regression, which are responsible for detecting the head, are often carried out in parallel, formulating a multi-task loss function for joint training. There is only one class prediction and position regression, and most of the weights are shared. Hence, one-stage algorithms are more time-efficient at the cost of accuracy. In any case, with the continuous development of deep learning in the field of computer vision, object detection algorithms are also constantly learning from and improving each other.

\subsection{Vision Transformer}
\begin{figure}[htbp]

  \centering
  % include second image
  \includegraphics[width=3.3in]{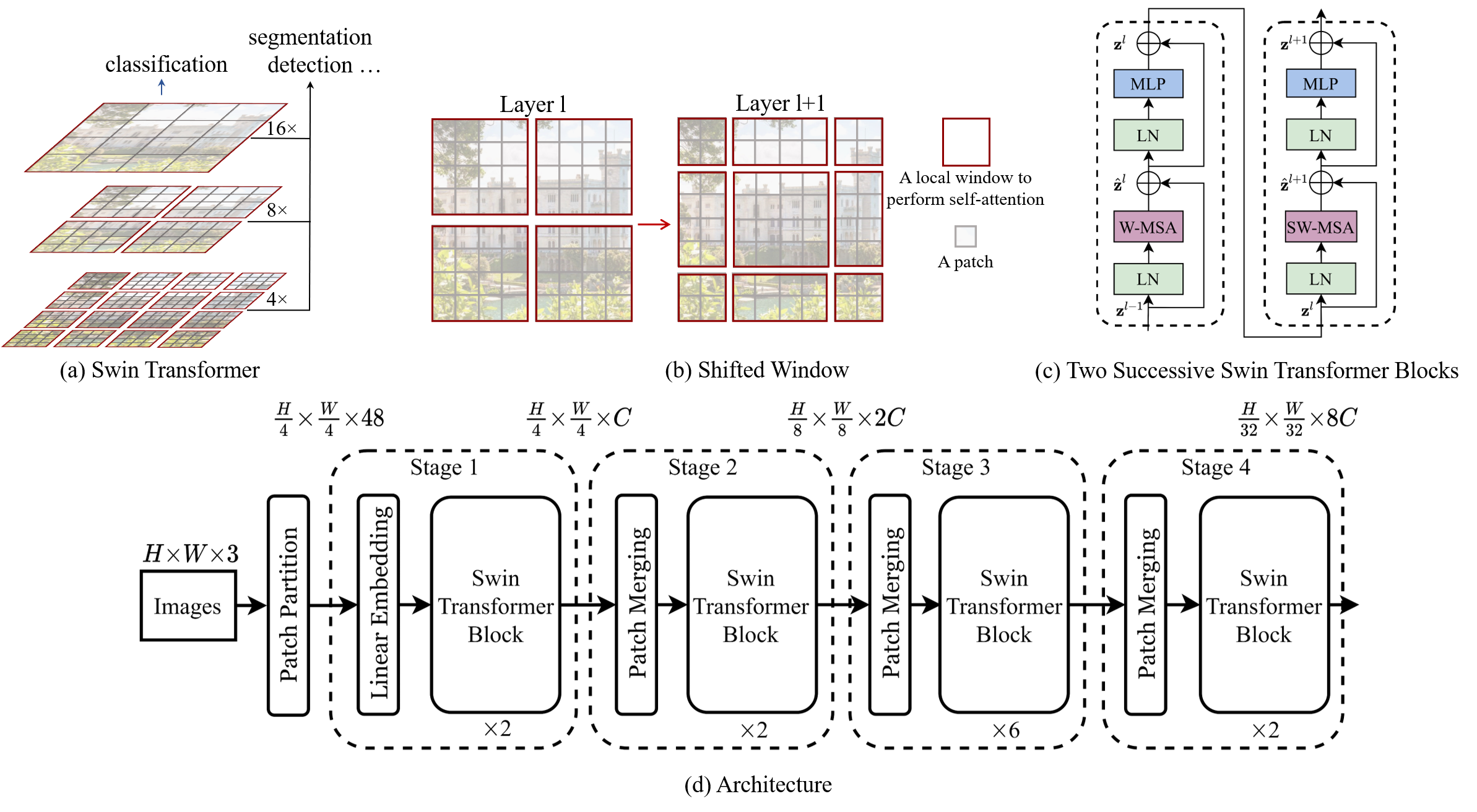}  
  \caption{Swin Transformer V2 framework \cite{liu2022swin}.}
  \label{swint2}

\end{figure}

ViTs have emerged as the most active research field in object detection tasks recently, with models like Swin-Transformer \cite{liu2021swin, liu2022swin}, ViTdet \cite{li2022exploring}, and DINO \cite{zhang2022dino} leading the forefront. Unlike conventional CNNs, ViTs leverage self-attention mechanisms to process image patches as sequences, offering a more flexible representation of spatial hierarchies. The core mechanism of these models involves dividing an image into a sequence of patches and applying Transformer encoders \cite{vaswani2017attention} to capture complex dependencies between them. This process enables ViTs to efficiently learn global context, which is pivotal in understanding comprehensive scene layouts and object relations. For instance, the Swin-Transformer \cite{liu2021swin} introduces a hierarchical structure with shifted windows, enhancing the model's ability to capture both local and global features. In the following, the Swin-Transformer was scaled to Swin-Transformer V2 \cite{liu2022swin} with the capability of training high-resolution images (see Fig. \ref{swint2}).

The primary advantages of ViTs in object detection are their scalability to large datasets and superior performance in capturing long-range dependencies. This makes them particularly effective in scenarios where contextual understanding is crucial. Additionally, ViTs demonstrate strong transfer learning capabilities, performing well across various domains with minimal fine-tuning. However, challenges with ViTs include their computational intensity due to self-attention mechanisms, particularly when processing high-resolution images. This can limit their deployment in real-time applications where computational resources are constrained. Additionally, ViTs often require large-scale datasets for pre-training to achieve optimal performance, which can be a limitation in data-scarce environments. Despite these challenges, ongoing advancements in ViT architectures, such as the development of efficient attention mechanisms \cite{shen2021efficient} and hybrid CNN-Transformer models \cite{maaz2022edgenext}, continue to enhance their applicability and performance in diverse object detection tasks.

\begin{figure}[!tb]

  \centering
  % include second image
  \includegraphics[width=.9\linewidth]{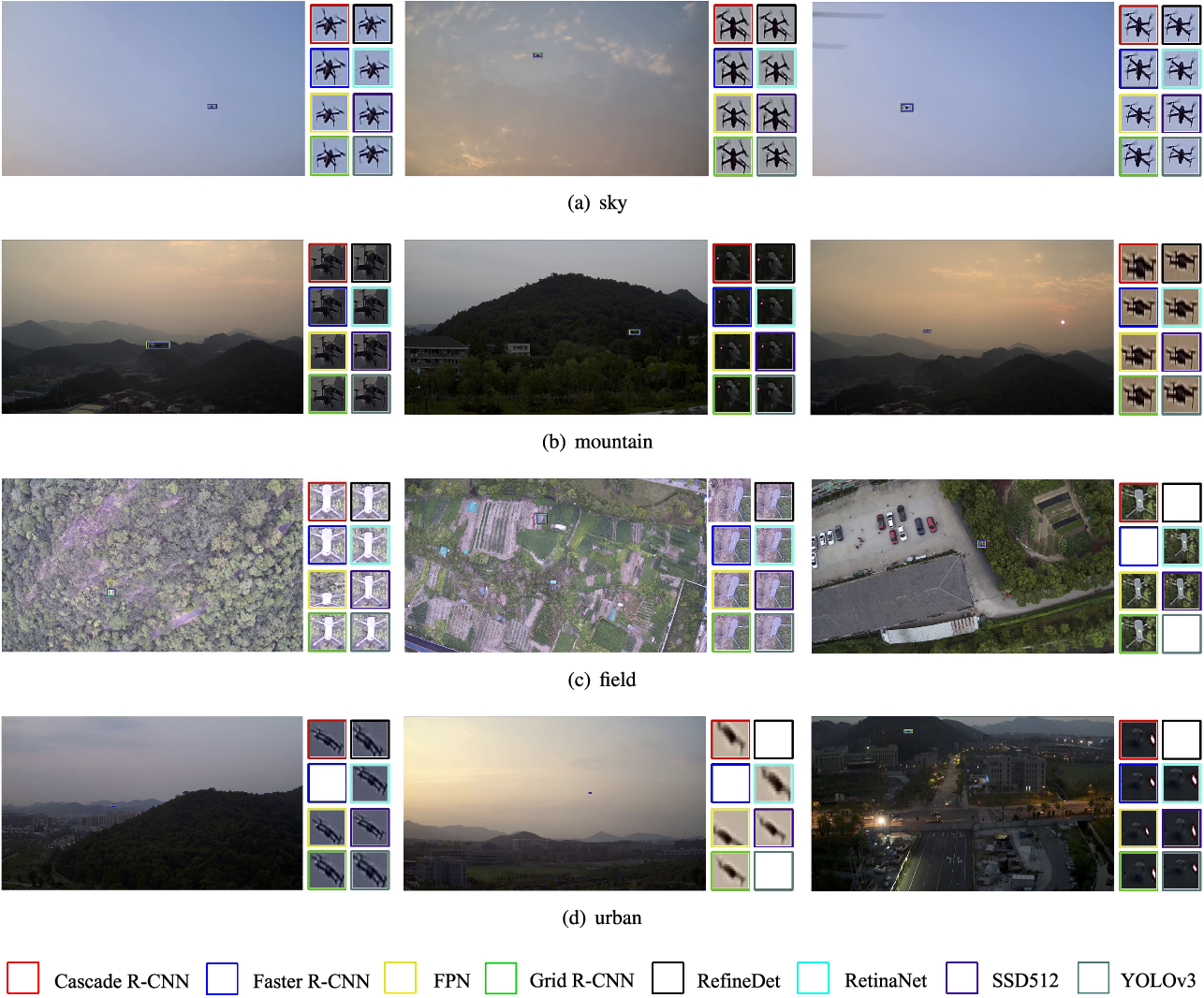}  
  \caption{Air-to-air object detection of micro-UAVs with a monocular camera \cite{zheng2021air}.}
  \label{air-to-air}

\end{figure}

When applying object detection algorithms to drone applications, it is necessary to find the best balance between computation speed and accuracy. Besides, massive drone datasets are required for training and testing. Zheng Ye \textit{et al.} \cite{zheng2021air} collected an air-to-air drone dataset ``Det-Fly" (see Fig. \ref{air-to-air}) and evaluated air-to-air object detection of a micro-UAV with eight different object detection algorithms, namely RetinaNet \cite{lin2017focal}, SSD, Faster R-CNN, YOLOv3 \cite{redmon2018yolov3}, FPN \cite{lin2017feature}, Cascade R-CNN\cite{cai2018cascade} and Grid R-CNN \cite{lu2019grid}. The evaluation results in \cite{zheng2021air} showed that the overall performance of Cascade R-CNN and Grid R-CNN is superior compared to the others. However, the YOLOv3 provides the fastest inference speed among others. Wei Xun \textit{et al.} \cite{9358449} conducted another investigation into drone detection, employing the YOLOv3 architecture and deploying the model on the NVIDIA Jetson TX2 platform. They collected a dataset comprising 1435 images featuring various UAVs, including drones, hexacopters, and quadcopters. Utilizing custom-trained weights, the YOLOv3 model demonstrated proficiency in drone detection within images. However, the deployment of this trained model faced constraints due to the limited computation capacity of the Jetson TX2, which posed challenges for effective real-time application.

\begin{figure*}[!tbp]
    \centering
    \includegraphics[width=6in]{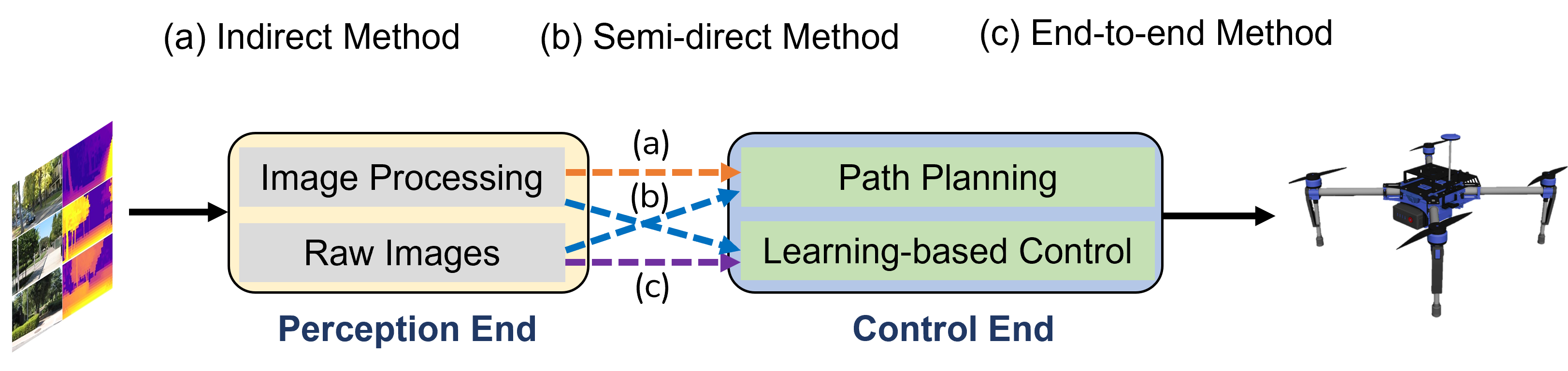}
    \caption{Vision-based control methods for drone applications. Based on the ways of visual perception and control, the methods can be divided into indirect methods, semi-direct methods, and end-to-end methods.}
    \label{methods}
\end{figure*}

In agile flight, computation speed is more important than accuracy since real-time object detection is required to avoid obstacles swiftly. Therefore, a simple gate detector and a filter algorithm are adopted as the basis of the visual perception of Swift \cite{kaufmann2023champion}. Considering the agility of the drone, a stabilization module is required to obtain more robust and accurate object detection and tracking results in real-time flight. Moreover, in the drone datasets covered in existing works, each image only includes a single UAV. To classify and detect different classes of drones in multi-drone systems, a new dataset of multiple types of drones has to be built from scratch. Furthermore, the dataset can be adapted to capture adversary drones in omnidirectional visual perception to enhance avoidance capability.

\section{Vision-based Control} \label{sec:vision-control}
Vision-based control for robotics has been widely studied in recent years, whether for ground robots or aerial robotics such as drones. For drones flying in a GPS-denied environment, visual inertial odometry (VIO) \cite{Aqel2016,Delmerico2018,Scaramuzza2020} and visual simultaneous localization and mapping systems (SLAM) \cite{mur2015orb,Qin2018, Campos2021} have been preferred choices for navigation. Meanwhile, in a clustered environment, research on obstacle avoidance \cite{Loquercio2021, Ross2013, xie2017towards, zhou2020ego} based on visual perception has attracted much attention in the past few years. Obstacle avoidance has been a main task for vision-based control as well as for the current learning algorithms of drones.

From the perspective of how drones obtain visual perception (perception end) and how drones generate control commands from visual perception (control end), existing vision-based control methods can be categorized into indirect methods, semi-direct methods, and end-to-end methods. The relationship between these three categories is illustrated in Fig. \ref{methods}. In the following, we will discuss and evaluate these methods in three different categories, respectively.

\subsection{Indirect Methods}
\begin{figure}[!tbp]
    \centering
    \includegraphics[width=3.3in]{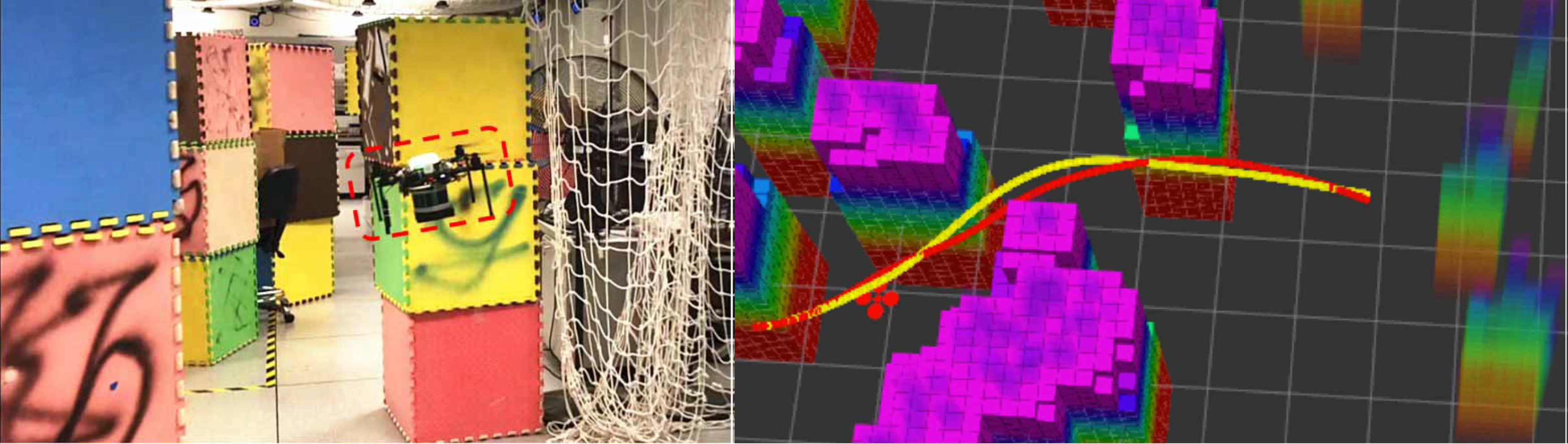}
    \caption{A vision-based drone is traversing through a cluttered indoor environment with generated point cloud maps and online path planning \cite{zhou2019robust}.}
    \label{indirect_method}
\end{figure}

Indirect methods \cite{mohta2018fast,falanga2020dynamic,Gao2018Optimal,gao2020teach, Gao2019Flying, gao2019, zhou2019robust, zhou2020ego, zhou2021raptor,Quan2021, zhang2022self} refer to extracting features from images or videos to generate visual odometry, depth maps, and 3D point cloud maps for drones to perform path planning based on traditional optimization algorithms (see Fig. \ref{indirect_method}). Obstacle states, such as 3D shape, position, and velocity, are detected and mapped before a maneuver is taken. Once online maps are built or obstacles are located, the drone can generate a feasible path or take actions to avoid obstacles.

\begin{figure}[!tbp]
    \centering
    \includegraphics[width=3.3in]{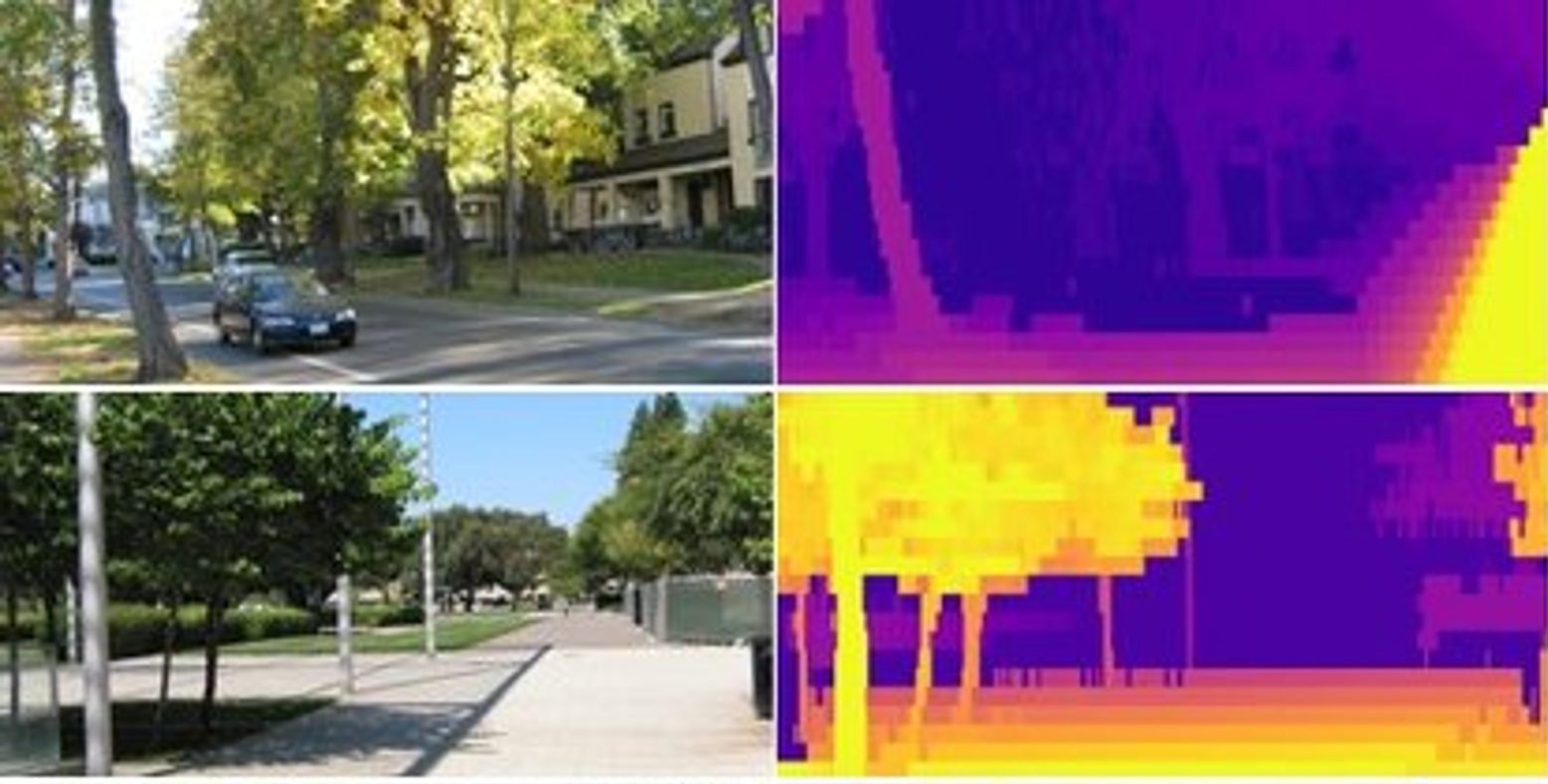}
    \caption{Depth maps generated from the viewpoint of a drone \cite{huang2019unsupervised}.}
    \label{depth_img}
\end{figure}

SOTA indirect methods generally divide the mission into several subtasks, namely \textbf{perception}, \textbf{mapping} and \textbf{planning}. On the perception side, depth images are always required to generate corresponding distance and position information for navigation. A depth image is a grey-level or color image that can represent the distance between the surfaces of objects from the viewpoint of the agent. Fig. \ref{depth_img} shows color images and corresponding depth images from a drone’s viewpoint. The illuminance is proportional to the distance from the camera. A lighter color denotes a nearer surface, and darker areas mean further surfaces. A depth map provides the necessary distance information for drones to make decisions to avoid static and dynamic obstacles. Currently, off-the-shelf RGB-D cameras, such as the Intel RealSense depth camera D415, the ZED 2 stereo camera, and the Structure Core depth camera, are widely used for drone applications. Therefore, traditional obstacle avoidance methods can treat depth information as a direct input. However, for omnidirectional perception in wide-view scenarios, efficient onboard monocular depth estimation is always required, which is a challenge to address with existing methods.

On the mapping side, a point cloud map \cite{rusu2008towards} or Octmap \cite{hornung13auro}, representing a set of data points in a 3D space is commonly generated. Each point has its own Cartesian coordinates and can be used to represent a 3D shape or an object. A 3D point cloud map is not from the view of a drone but constructs a global 3D map that provides global environmental information for a drone to fly. The point-cloud map can be generated from a LIDAR scanner or many overlapped images combined with depth information. An illustration of an original scene and a point cloud map are shown in Fig. \ref{indirect_method}, where the drone can travel around without colliding with static obstacles.

Planning is a basic requirement for a vision-based drone to avoid obstacles. Within the indirect methods, planning can be further divided into two categories: one is offline methods based on high-resolution maps and pre-known position information, such as Dijkstra's algorithm \cite{Dijkstra1959ANO}, A-star \cite{4082128}, RRT-connect\cite{kuffner2000rrt} and sequential convex optimization \cite{augugliaro2012generation}; the other is online methods based on real-time visual perception and decision-making. Online methods can be further categorized into online path planning \cite{zhou2019robust, zhou2020ego, zhou2021raptor} and artificial potential field (APF) methods \cite{falanga2020dynamic, iswanto2019artificial}.

Most vision-based drones rely on online methods. Compared to offline methods, which require an accurate pre-built global map, online methods provide advanced maneuvering capabilities for drones, especially in a dynamic environment. Currently, due to the advantages of optimization and prediction capabilities, online path planning methods have become the preferred choice for drone obstacle avoidance. For instance, in the SOTA work \cite{zhou2019robust}, Zhou Boyu \textit{et al.} introduced a robust and efficient motion planning system called Fast-Planner for a vision-based drone to perform high-speed flight in an unknown cluttered environment. The key contributions of this work are a robust and efficient planning scheme incorporating path searching, B-spline optimization, and time adjustment to generate feasible and safe trajectories for vision-based drones' obstacle avoidance. Using only onboard vision-based perception and computing, this work demonstrated agile drone navigation in unexplored indoor and outdoor environments. However, this approach can only achieve maximum speeds of $3m/s$ and requires $7.3ms$ for computation in each step. To improve the flight performance and save computation time, Zhou Xin \textit{et al.} \cite{zhou2020ego} provided a Euclidean Signed Distance Field (ESDF)-free gradient-based planning framework solution, EGO-Planner, for drone autonomous navigation in unknown obstacle-rich situations. Compared to the Fast-Planner, the EGO-Planner achieved faster speeds and saved a lot of computation time. However, these online path planning methods require bulky visual sensors, such as RGBD cameras or LIDAR, and a powerful onboard computer for the complex numerical calculation to obtain a local or global optimal trajectory.

In contrast to online path planning, the artificial potential field methods require less computation resources and can well cope with dynamic obstacle avoidance using limited sensor information. The APF algorithm is one of the algorithms in the robot path planning approach that uses attractive force to achieve the objective position and repulsive force to avoid obstacles in an unknown environment \cite{khatib1985real}. Falanga \textit{et al.} \cite{falanga2020dynamic} developed an efficient and fast control strategy based on the artificial potential field method to avoid fast approaching dynamic obstacles. The obstacles in \cite{falanga2020dynamic} are represented as repulsive fields that decay over time, and the repulsive forces are generated from the first-order derivation of the repulsive fields at each time step. However, the repulsive forces computed only reach substantial values when the obstacle is very close, which may lead to unstable and aggressive behavior. Besides, artificial potential field methods are heuristic methods that cannot guarantee global optimization and robustness for drones. Hence, it is not ideal to adopt potential field methods to navigate through cluttered environments.

\begin{figure}[!tbp]
\centering
\begin{subfigure}{0.24\textwidth}
    \centering
  % include first image
    \centering
    \includegraphics[height=1\linewidth]{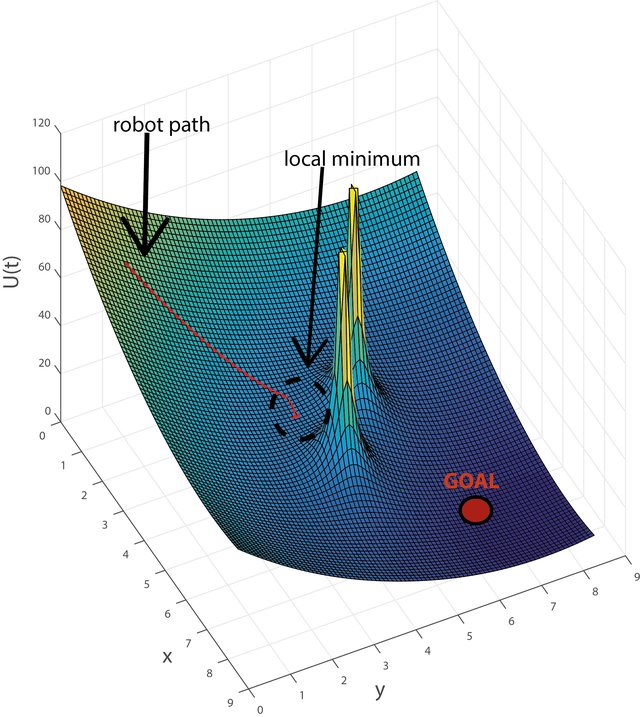}
    \caption{}
\end{subfigure}
\begin{subfigure}{0.24\textwidth}
  \centering
  % include second image
  \includegraphics[height=1\linewidth]{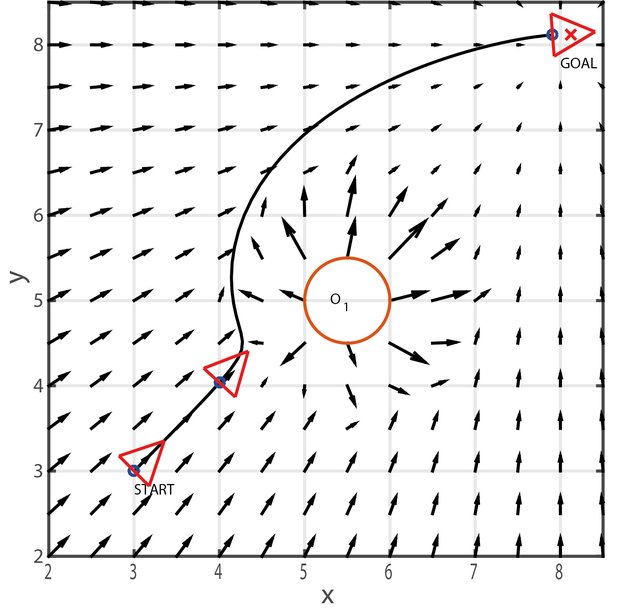}
  \caption{}
\end{subfigure}
\caption{Artificial potential field (APF) methods in drones' obstacle avoidance \cite{iswanto2019artificial} (a) Generated artificial potential field; (b) Flight trajectory based on the APF.}
\label{apf}
\end{figure}

\subsection{End-to-end Methods}
\begin{figure}[htbp]
\centering
  % include second image
 \includegraphics[width=0.8\linewidth]{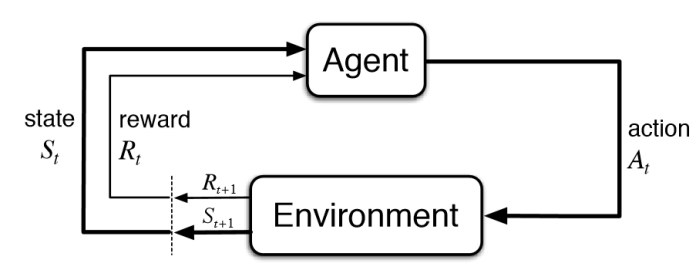}
\caption{The reinforcement learning process.}
\label{rl}
\end{figure}

In contrast to the indirect methods, which divide the whole mission into multiple sub-tasks, such as perception, mapping, and planning, end-to-end methods \cite{zhu2017target,dai2020automatic, anwar2018navren, Loquercio2021, zhang2023partially} combine computer vision and reinforcement learning (RL) to map the visual observations to actions directly. RL \cite{sutton2018reinforcement} is a technique for mapping the state (observation) space to the action space in order to maximize a long-term return with given rewards. The learner is not explicitly told what action to carry out but must figure out which action will yield the highest reward during the exploring process. A typical RL model (see Fig. \ref{rl}) features agents, the environment, reward functions, action, and state space. The policy model achieves convergent status via constant interactions between the agents and the environment, where the reward function guides the training process.

\begin{figure}[!tbp]
\centering
\begin{subfigure}{0.23\textwidth}
    \centering
  % include first image
    \centering
    \includegraphics[height=0.6\linewidth]{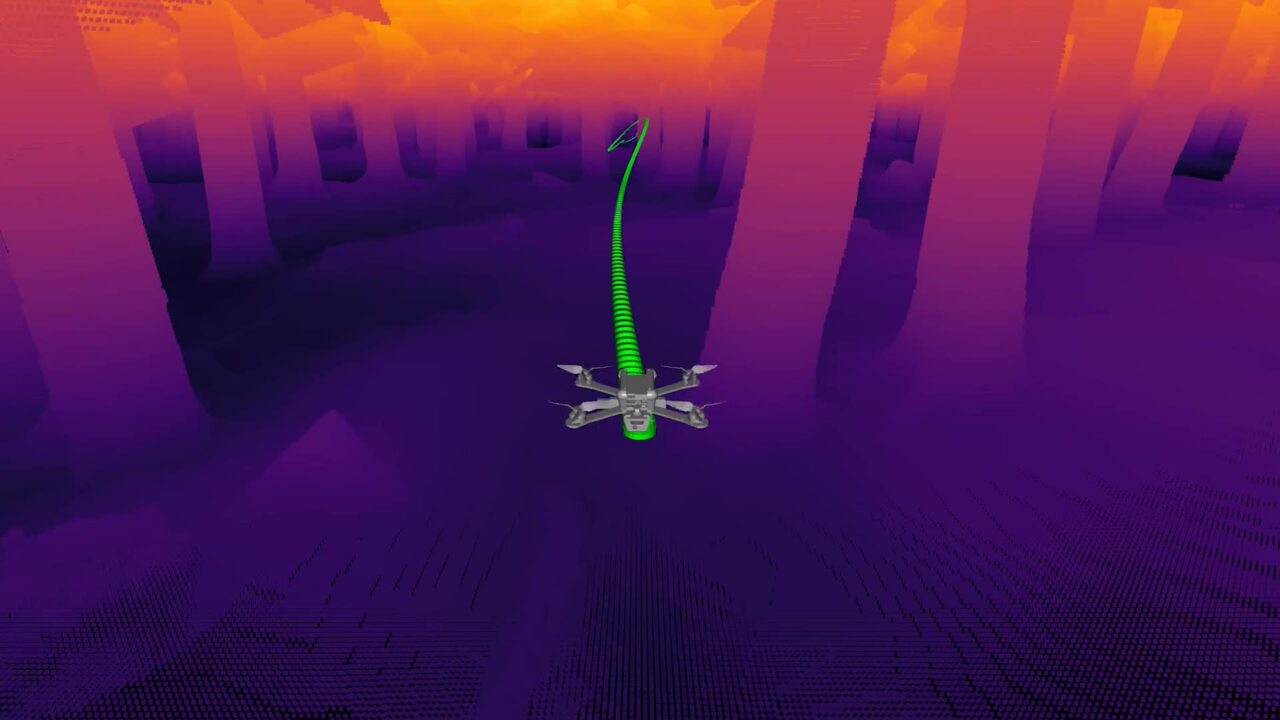}
    \caption{}
\end{subfigure}
\begin{subfigure}{0.23\textwidth}
  \centering
  % include second image
  \includegraphics[height=0.6\linewidth]{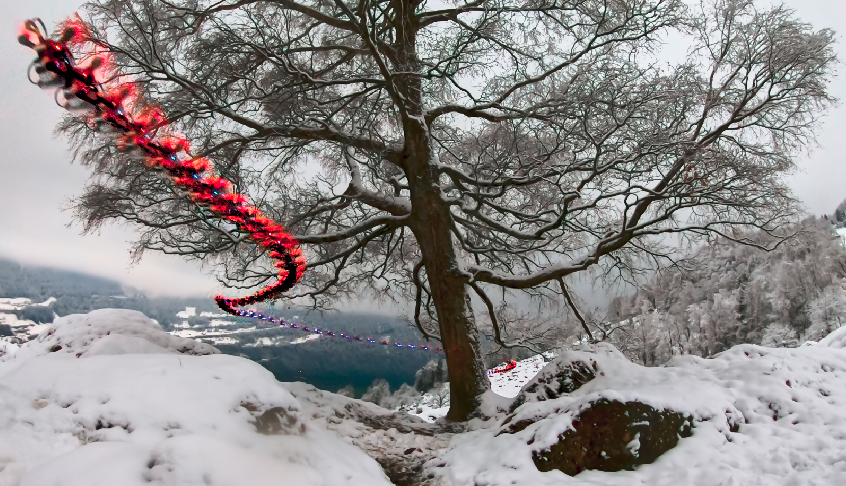}
  \caption{}
\end{subfigure}
\caption{Quadrotor drone flying in a wild environment with end-to-end reinforcement learning \cite{Loquercio2021}. (a) Training process in the simulation platform; (b) Real flight test in the wild snow environment.}
\label{drl_wild}
\end{figure}

With end-to-end methods, the visual perception of drones is encoded by a deep neural network into an observation vector of the policy network. The mapping process is usually trained offline with abundant data, which requires high-performance computers and simulation platforms. The training data set is collected from expert flight demonstrations for imitation learning or from simulations for online training. To better generalize the performance of a trained neural network model, scenario randomization (domain randomization) is essential during the training process. Malik Aqeel Anwar \textit{et al.} \cite{anwar2018navren} presented an end-to-end reinforcement learning approach called NAVREN-RL to navigate a quadrotor drone in an indoor environment with expert data and knowledge-based data aggregation. The reward function in \cite{anwar2018navren} was formulated from a ground truth depth image and a generated depth image. Loquercio \textit{et al.} \cite{Loquercio2021} developed an end-to-end approach that can autonomously guide a quadrotor drone through complex wild and human-made environments at high speeds with purely onboard visual perception (depth image) and computation. The neural network policy was trained in a high-fidelity simulation environment with massive expert knowledge data. While end-to-end methods provide us with a straightforward way to generate obstacle avoidance policies for drones, they require massive training data with domain randomization (usually counted in the millions) to obtain acceptable generalization capabilities. Meanwhile, without expert knowledge data, it is challenging for the neural network policy to update its weights when the reward space is sparse. To implement end-to-end methods, we commonly consider the following aspects: neural network architecture, training process, and Sim2Real transfer.

\begin{figure}[!tbp]
\centering
\begin{subfigure}{0.46\textwidth}
    \centering
  % include first image
    \centering
    \includegraphics[height=0.7\linewidth]{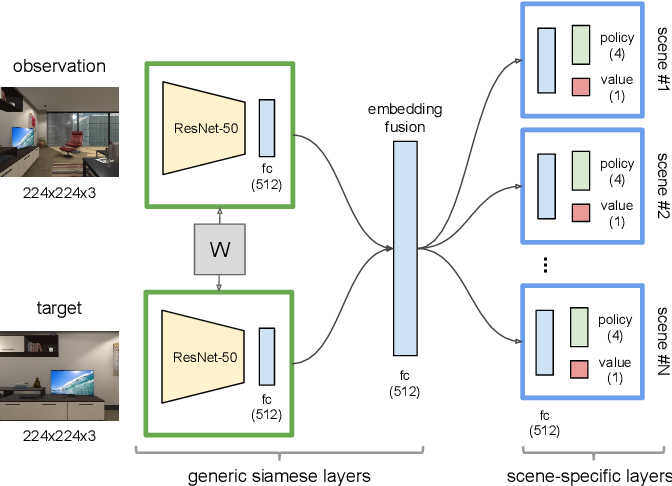}
    \caption{}
\end{subfigure}
\begin{subfigure}{0.46\textwidth}
  \centering
  % include second image
  \includegraphics[height=0.7\linewidth]{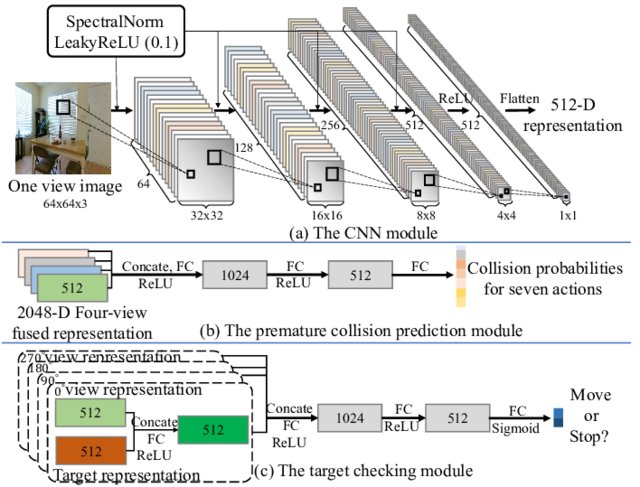}
  \caption{}
\end{subfigure}
\caption{The neural network architectures of end-to-end reinforcement learning methods. (a) Target driven navigation with ResNet50 encoder\cite{zhu2017target}; (b) Target driven visual navigation for a  robot with shallow CNN encoder \cite{Wu2021}.}
\label{nn-arch}
\end{figure}

\textbf{Neural Network Architecture:} The neural network architecture is the core component in the end-to-end methods, which determines the computation efficiency and intelligence level of the policy. In the end-to-end method, the input of the neural network architecture is the image raw data (RGB/RGBD), and the output is the action vector an agent needs to take. The images are encoded into a vector and then concatenated with other normalized observations to form an input vector of a policy network. For the image encoder, there are many pre-trained neural network architectures that can be considered, such as ResNet \cite{he2016deep}, VGG \cite{simonyan2014very}, and nature Convolutional Neural Network (CNN) \cite{Mnih2015}. Zhu \textit{et al.} \cite{zhu2017target} developed a target-driven visual navigation approach for robots with end-to-end deep reinforcement learning, where a pre-trained ResNet-50 is used to encode the image into a feature vector. In \cite{Wu2021}, a more data-efficient image encoder with 5 layers was designed for target-driven visual navigation for robots with end-to-end imitation learning. Before designing the neural network architecture, we first need to determine the observation space and action space of the task. The training efficiency and space complexity are the two aspects we need to consider in the designing process.

\textbf{Training Process:} The training process is the most time-consuming part of the end-to-end learning methods. It requires the use of gradient information from the loss function to update the weights of the neural network and improve the policy model. There are two main training algorithms for reinforcement learning, namely the on-policy algorithm and the off-policy algorithm. On-policy algorithms attempt to evaluate or improve the policy that is used to make decisions, i.e., the behavior policy for generating actions is the same as the target policy for learning. When the agent is exploring the environment, on-policy algorithms are more stable than off-policy algorithms when the agent is exploring the environment. SARSA (State-Action-Reward-State-Action experiences to update the Q-values) \cite{sutton2018reinforcement} is an on-policy reinforcement learning algorithm that estimates the value function of the policy being carried out. PPO \cite{schulman2017proximal} is another efficient on-policy algorithm widely used in reinforcement learning. By compensating for the fact that more probable actions are going to be taken more often, PPO addresses the issue of high variance and low convergence with policy gradient methods. To avoid large weights' vibration, a “clipped”
policy gradient update formulation is designed. 

\begin{figure}[!tbp]
\centering
  % include second image
 \includegraphics[height=0.5\linewidth]{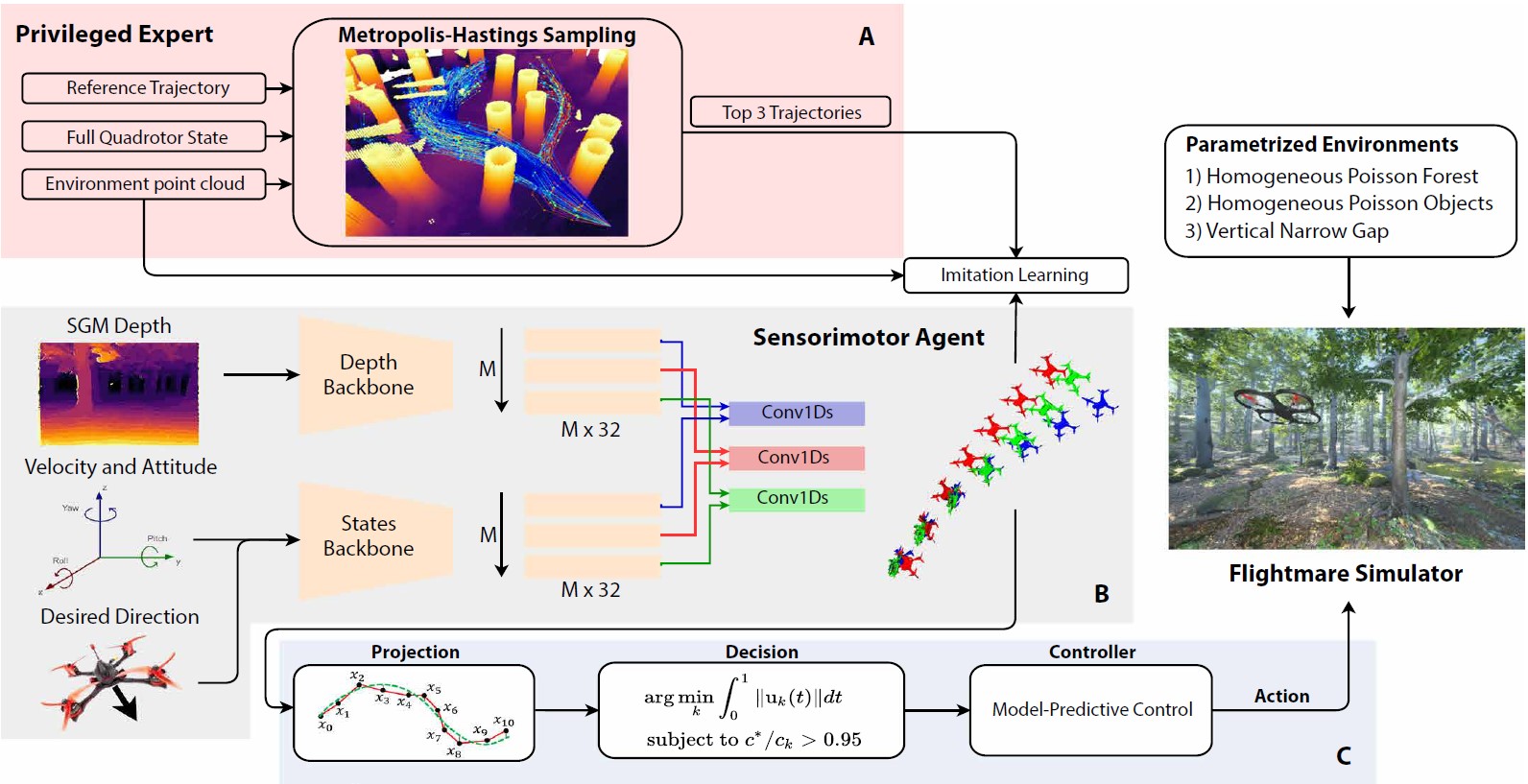}
\caption{Training process used in \cite{Loquercio2021} to fly an agile drone through forest.}
\label{training}
\end{figure}

In contrast, off-policy algorithms evaluate or improve a policy different from that used to generate the action currently. The off-policy algorithms such as Q-learning \cite{watkins1992q,van2016deep} try to learn a greedy policy in every step. Off-policy algorithms perform better at movement predictions, especially in an unknown environment. However, off-policy algorithms can be very unstable due to the unstatic environments the offline data seldom covers. Imitation learning \cite{zhu2020off} is another form of off-policy algorithm that tries to mimic demonstrated behavior in a given task. Through training from demonstrations, imitation learning can save remarkable exploration costs. \cite{Loquercio2021} adopted imitation learning to speed up the training process. The whole training process is illustrated in Fig. \ref{training}. With this privileged expert knowledge, the policy can be trained to find a time-efficient trajectory to avoid obstacles.

For multi-agent systems, multi-agent reinforcement learning (MARL) is widely studied to encourage agent collaboration. In MARL, the credit assignment is a crucial issue that determines the contribution of each agent to the group's success or failure. COMA\cite{foerster2018counterfactual} is a baseline method that uses a centralized critic to estimate the action-value function, and for each agent, it computes a counterfactual advantage function to represent the value difference, which can determine the contribution of each agent. However, COMA still does not fully address the model complexity issue. QMIX \cite{rashid2020monotonic} is well developed to address the scalability issue by decomposing the global action-value function into individual agent's value functions. However, QMIX assumes the environment is fully observable and may not be able to handle scenarios with continuous action space. Hence, attention mechanism-enabled MARL is a promising direction to address the variable observations. Besides, to balance individual and team reward, a MARL with mixed credit assignment algorithm, POCA-Mix, was proposed in \cite{10195084} to achieve collaborative multi-target search with a visual drone swarm.

\textbf{Sim2Real Transfer:} For complex tasks, the policy neural networks are usually trained on simulation platforms such as AirSim \cite{shah2018airsim}, Unity \cite{juliani2018unity} or Gazebo \cite{zamora2016extending}. The differences between the simulation and the real environment are non-negligible. Sim2Real is the way to deploy the neural network model trained in a simulation environment on a real physical agent. For deployment, it is essential to validate the generalization capability of trained neural network models. A wide variety of Sim2Real techniques \cite{ christiano2016transfer,tan2018sim, finn2017model, tobin2017domain, andrychowicz2020learning} have been developed to improve the generalization and transfer capabilities of models. Domain randomization \cite{tobin2017domain} is one of these techniques to improve the generalization capability of the trained neural network to unseen environments. Domain randomization is a method of trying to discover a representation that can be used in a variety of scenes or domains. Existing domain randomization techniques \cite{xiao2021flying, Loquercio2021} for drones' obstacle avoidance include position randomization, depth image noise randomization, texture randomization, size randomization, etc. Therefore, we need to apply domain randomization in our training process to enhance the generalization capability of the trained model in deployment.

\begin{figure}[!tbp]
\centering
  % include second image
 \includegraphics[height=0.5\linewidth]{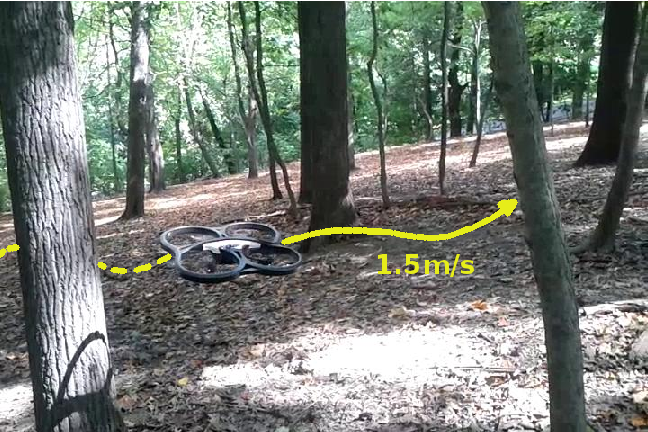}
\caption{A drone avoids the obstacles in the forest with semi-direct methods \cite{Ross2013}.}
\label{semi-direct}
\end{figure}

\subsection{Semi-direct Methods}
Compared to the end-to-end methods, which generate actions from image raw data directly, semi-direct methods \cite{Ross2013,Akhloufi2019,geyer20063d,kumar2021omnidet} introduce an intermediate phase for drones to take actions from visual perception, aiming to improve the generalization and transfer capabilities of the methods over unseen environments. There are two ways to design the semi-direct method architectures: one is to generate the required information from image processing (such as the relative positions of obstacles from object detection and tracking or the point cloud map) and train the control policy with deep reinforcement learning; another is to obtain the required states (such as depth image) directly from the raw image data with deep learning and avoid the obstacles using numerical or heuristic methods. These two methods can be denoted as indirect (front end)-direct (back end) methods and direct (front end)-indirect (back end) methods.

\begin{table*}[!tb]
\centering
\caption{Summary of Vision-Based Control Methods in Drone Technology}
\begin{tabular}{|l|p{0.15\textwidth}|p{0.15\textwidth}|p{0.1\textwidth}|p{0.35\textwidth}|}
\hline
\textbf{Method Type} & \textbf{Perception End} & \textbf{Control End} & \textbf{Key Studies} & \textbf{Description and Applicability} \\
\hline
\multirow{2}{*}{Indirect Methods} & Depth maps, 3D point cloud maps & Traditional optimization algorithms & \cite{mohta2018fast}, \cite{Gao2018Optimal}, \cite{gao2020teach}, \cite{zhou2021raptor} & Focus on generating visual odometry, depth maps, and 3D point cloud maps for path planning. Suitable for safety-critical tasks with accurate models.  \\
\cline{2-5}
 & Depth maps, 3D point cloud maps & Online path planning, APF, etc.  & \cite{falanga2020dynamic}, \cite{Quan2021} & Utilize depth information for real-time visual perception and decision-making in dynamic environments. Suitable for rapid response tasks with certain information. \\
\hline
\multirow{2}{*}{End-to-End Methods} & Visual observations encoded by DNNs & RL for action mapping & \cite{xiao2022collaborative}, \cite{dai2020automatic}, \cite{anwar2018navren} & Combine deep learning for visual perception with RL for direct action response. Suitable for complex tasks with uncertain information. \\
\cline{2-5}
 & Encoded visual observations & Imitation learning and online training & \cite{8798720}, \cite{Loquercio2021} & Use deep learning for visual encoding and train the control system with expert demonstrations and online training. Suitable for sample-efficient tasks with expert demonstrations. \\
\hline
\multirow{2}{*}{Semi-Direct Methods} & Intermediate features from image processing & DRL for action & \cite{Ross2013}, \cite{Akhloufi2019}, \cite{kaufmann2023champion}, \cite{10184073} & Extract intermediate features like relative positions or velocities and use DRL for action decisions. Suitable for complex tasks with high generalization requirement. \\
\cline{2-5}
 & Raw image data for depth images or obstacle tracking & Numerical/Heuristic methods for obstacle avoidance & \cite{geyer20063d}, \cite{kumar2021omnidet} & Utilize direct image data to obtain necessary states for obstacle avoidance using non-learning-based methods. Suitable for tasks where robust stereo information is not available. \\
\hline
\end{tabular}
\label{table:vision_control_methods}
\end{table*}

\textbf{Indirect-direct methods} \cite{Ross2013,Akhloufi2019} firstly obtain intermediate features such as relative position or velocity of the obstacles from image processing and then use this intermediate feature information as observations to train the policy neural network via deep reinforcement learning. Indirect-direct methods generally rely on designing suitable intermediate features. In \cite{Ross2013}, the features related to depth cues such as Radon features (30 dimensional), structure tensor statistics (15 dimensional), Laws' masks (8 dimensional), and optical flow (5 dimensional) were extracted and concatenated into a single feature vector as visual observation. Together with the other nine additional features, the control policy was trained with imitation learning to navigate the drone through a dense forest environment (see Fig. \ref{semi-direct}).

\begin{figure}[!tbp]
\centering
  % include second image
 \includegraphics[width=3.3in]{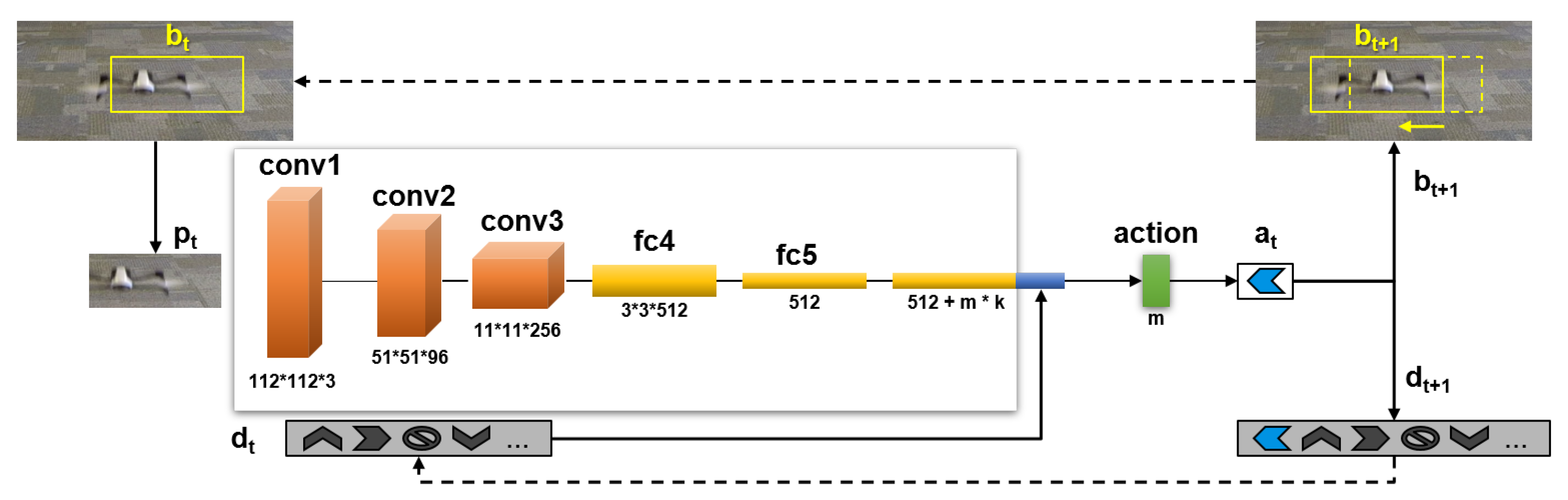}
\caption{A drone is tracking and chasing a target drone with object detection and deep reinforcement learning \cite{Akhloufi2019}.}
\label{drone-chase}
\end{figure}

Moulay \textit{et al.} \cite{Akhloufi2019} proposed a semi-direct vision-based learning control policy for UAV pursuit-evasion. Firstly, a deep object detector (YOLOv2) and a search area proposal (SAP) were used to predict the relative position of the target UAV in the next frame for target tracking. Afterward, deep reinforcement learning (see Fig. \ref{drone-chase}) was adopted to predict the actions the follower UAV needs to perform to track the target UAV. Indirect-direct methods are able to improve the generalization capability of policy neural networks, but at the cost of heavy computation and time overhead.

\textbf{Direct-indirect methods} try to obtain depth images \cite{godard2017unsupervised, godard2019digging} and track obstacles \cite{liu2021yolostereo3d, mancini2016fast} from training and use non-learning-based methods, such as path planning or APF to avoid obstacles. Direct-indirect methods can be applied to a microlight drone with only monocular vision, but they require a lot of training data to obtain depth images or 3D poses of the obstacles. Michele \textit{et al.} \cite{mancini2016fast} developed an object detection system to detect obstacles at a very long range and at a very high speed, without certain assumptions on the type of motion. With a deep neural network trained on real and synthetic image data, fast, robust and consistent depth information can be used for drones' obstacle avoidance. Direct-indirect methods address the ego drift problem of monocular depth estimation using Structure from Motion (SfM) and provide a direct way to get depth information from the image. However, the massive train dataset and limited generalization capability are the main challenges for their further applications.

\begin{figure}[!tbp]
\centering
  % include second image
 \includegraphics[width=3.3in]{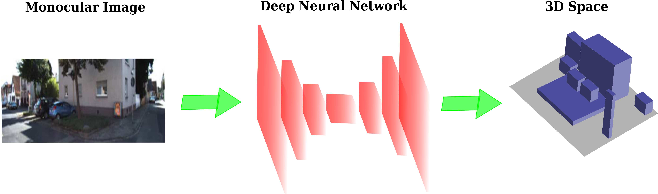}
\caption{Via deep learning, a 3D space generated from the monocular image for obstacle avoidance \cite{mancini2016fast}.}
\label{deep-depth}
\end{figure}

In summary, the field of vision-based control for drones encompasses a variety of methods, each with its own unique approach to perception and control. Indirect methods rely on traditional optimization algorithms and depth or 3D point cloud maps for navigation and obstacle avoidance. End-to-end methods leverage deep neural networks for visual perception and utilize reinforcement learning for direct action mapping. Semi-direct methods balance between computational efficiency and generalization by using intermediate features from image processing and a combination of DRL and heuristic methods for action generation. A comprehensive overview of these methods, along with key studies in each category, is summarized in Table \ref{table:vision_control_methods}, which provides a detailed comparison of their perception and control strategies.

\section{Applications and Challenges} \label{sec:app}
\subsection{Single Drone Application}
The versatility of single drones is increasingly recognized in a variety of challenging environments. These unmanned vehicles, with their inherent advantages, are effectively employed in critical areas such as hazardous environment detection and search and rescue operations. Single drone applications in vision-based learning primarily involve tasks like obstacle avoidance \cite{sanket2020evdodgenet, Loquercio2021}, surveillance \cite{singh2018eye, li2019simultaneously, zhou2016detecting, motlagh2017uav}, search-and-rescue operations \cite{goodrich2008supporting, lygouras2019unsupervised, tomic2012toward}, environmental monitoring \cite{senthilnath2017application, khosravi2021bl, donmez2021computer, lu2017species}, industrial inspection \cite{kim2019remote, khuc2020swaying, spencer2019advances} and autonomous racing \cite{li2020visual, muller2019learning, muller2018teaching}. Each field, while benefiting from the unique capabilities of drones, also presents its own set of challenges and areas for development.

\subsubsection{Obstacle Avoidance} The development of obstacle avoidance capabilities in drones, especially for vision-based control systems, poses significant challenges. Recent studies have primarily focused on static or simple dynamic environments, where obstacle paths are predictable \cite{kaufmann2018deep, Loquercio2021, Penicka2022}. However, complex scenarios involving unpredictable physical attacks from birds or intelligent adversaries remain largely unaddressed. For instance, \cite{kaufmann2018deep, Kaufmann2019, sanket2020evdodgenet, falanga2020dynamic} have explored basic dynamic obstacle avoidance but do not account for adversarial environments. To effectively handle such threats, drones require advanced features like omnidirectional visual perception and agile maneuvering capabilities. Current research, however, is limited in addressing these needs, underscoring the necessity for further development in drone technology to enhance evasion strategies against smart, unpredictable adversaries.

\subsubsection{Surveillance}While drones play a pivotal role in surveillance tasks, their deployment is not without challenges. Key obstacles include managing high data processing loads and addressing the limitations of onboard computational resources. In addressing these challenges, the study by Singh \textit{et al.} \cite{singh2018eye} presented a real-time drone surveillance system used to identify violent individuals in public areas. The proposed study was facilitated by cloud processing of drone images to address the challenge of slow and memory-intensive computations while still maintaining onboard short-term navigation capabilities. Additionally, in the study \cite{motlagh2017uav}, a drone-based crowd surveillance system was tested to achieve the goal of saving scarce energy of the drone battery. This approach involved offloading video data processing from the drones by employing the Mobile Edge Computing (MEC) method. Nevertheless, while off-board processing diminishes computational demands and energy consumption, it inevitably heightens the need for data transmission. Addressing the challenge of achieving real-time surveillance in environments with limited signal connectivity is an additional critical issue that requires resolution.

\subsubsection{Search and Rescue}In the field of search and rescue operations, a number of challenges exist that hinder the development of drone technology. A primary challenge faced by drones is extracting maximum useful information from limited data sources. This is crucial for improving the efficiency and success rate of these missions. Goodrich \textit{et al.} \cite{goodrich2008supporting} address this by developing a contour search algorithm designed to optimize video data analysis, enhancing the capability to identify key elements swiftly. However, incorporating temporal information into this algorithm introduces additional computational demands. These increased requirements present new challenges, such as the need for more powerful processing capabilities and potentially greater energy consumption. 

\subsubsection{Environmental Monitoring}A major challenge in the application of drones for environmental monitoring lies in efficiently collecting high-resolution data while navigating the constraints of battery life, flight duration, and diverse weather conditions. Addressing this, Senthilnath \textit{et al.} \cite{senthilnath2017application} showcased the use of fixed-wing and VTOL (Vertical Take-Off and Landing) drones in vegetation analysis, focusing on the challenge of detailed mapping through spectral-spatial classification methods. In another study, Lu \textit{et al.} \cite{lu2017species} demonstrated the utility of drones for species classification in grasslands, contributing to the development of methodologies for drone-acquired imagery processing, which is crucial for environmental assessment and management. While these studies represent significant steps in drone applications for environmental monitoring, several challenges persist. Future research may need to address the problems of improving the drones' resilience to diverse environmental conditions, and extend their operational range and duration to comprehensively cover extensive and varied landscapes.

\begin{figure}[!tb]
    \centering
    \begin{subfigure}{0.23\textwidth}
        \centering
        \includegraphics[width=0.93\textwidth, height=0.9\linewidth]{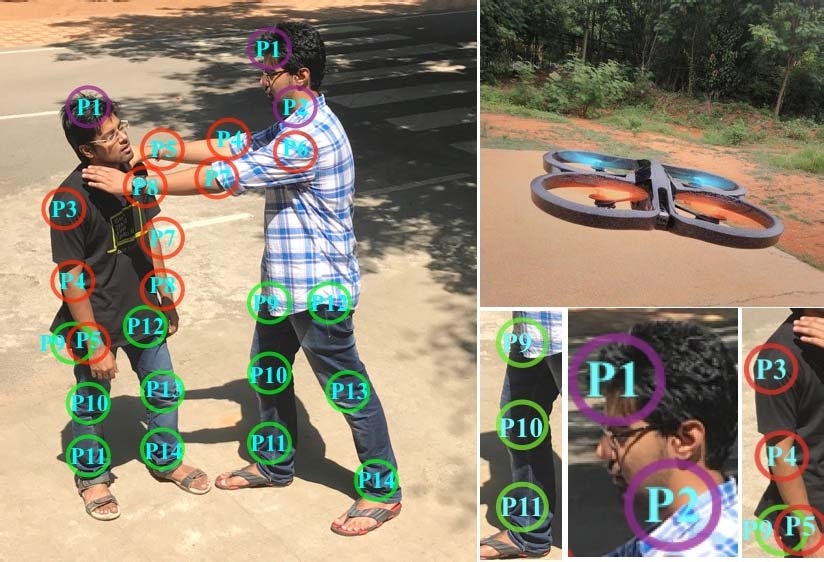}
        \caption{}
    \end{subfigure}%
    \begin{subfigure}{0.23\textwidth}
        \centering
        \includegraphics[width=0.93\textwidth, height=0.9\linewidth]{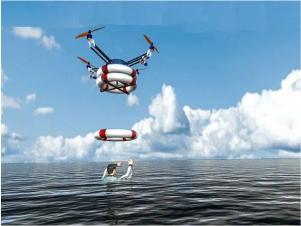}
        \caption{}
    \end{subfigure}
    \newline
    \begin{subfigure}{0.23\textwidth}
        \centering
        \includegraphics[width=0.93\textwidth, height=0.9\linewidth]{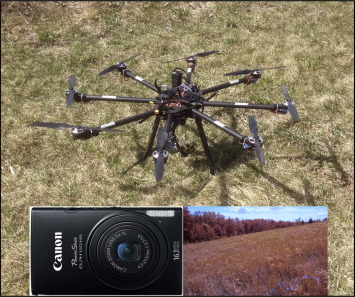}
        \caption{}
    \end{subfigure}%
    \begin{subfigure}{0.23\textwidth}
        \centering
        \includegraphics[width=0.93\textwidth, height=0.9\linewidth]{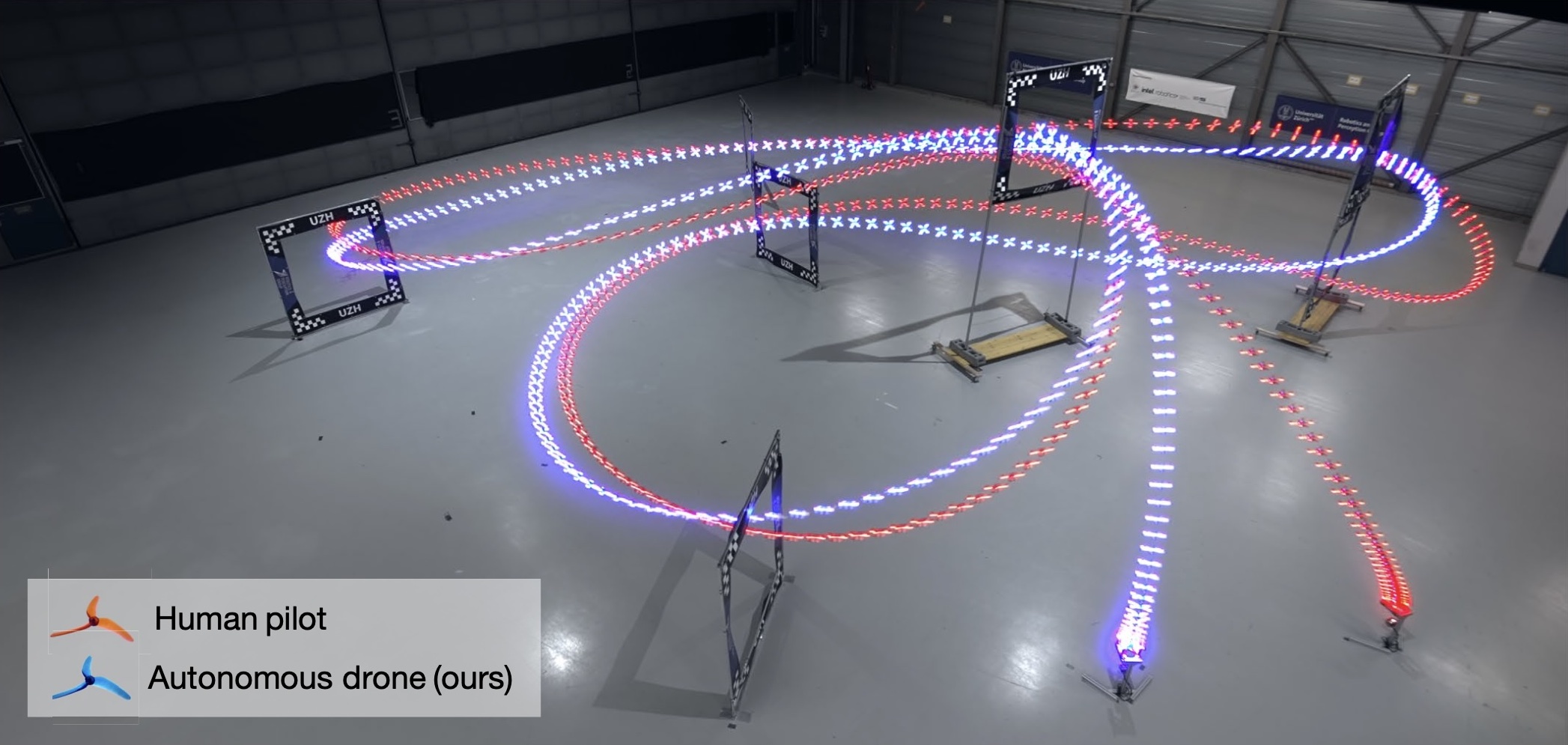}
        \caption{}
    \end{subfigure}
    \caption{Applications of single drone. (a) Surveillance \cite{singh2018eye}; (b) Search and Rescue \cite{xing2022multi}; (c) Environmental Monitoring \cite{lu2017species}; (d) Autonomous Racing \cite{kaufmann2023champion}.}
    \label{fig:single_applications}
\end{figure}

\subsubsection{Industrial Inspection}In industrial inspection, drones face key challenges like safely navigating complex environments and conducting precise measurements in the presence of various disturbances. Kim \textit{et al.} \cite{kim2019remote} addressed the challenge of autonomous navigation by using drones for proximity measurement among construction entities, enhancing safety in the construction industry. Additionally, Khuc \textit{et al.} \cite{khuc2020swaying} focused on precise structural health inspection with drones, especially in high or inaccessible locations. Despite these advancements in autonomous navigation and measurement accuracy, maintaining data accuracy and reliability in industrial settings with interference from machinery, electromagnetic fields, and physical obstacles continues to be a significant challenge, necessitating further research and development in this domain.

\subsubsection{Autonomous Racing}In autonomous drone racing, the central challenge is reducing delays exist in visual information processing and decision making and enhancing the adaptability of perception networks. In \cite{li2020visual}, a novel sensor fusion method was proposed to enable high-speed autonomous racing for mini-drones. This work also addressed issues with occasional large outliers and vision delays commonly encountered in fast drone racing. Another work \cite{muller2019learning} introduced an innovative approach to drone control, where a deep neural network (DNN) was used to fuse trajectories from multiple controllers. In the latest work \cite{kaufmann2023champion}, the vision-based drone outperformed world champions in the racing task, purely relying on onboard perception and a trained neural network. The primary challenge in autonomous drone racing, as identified in these studies, lies in the need for improved adaptability of perception networks to various environments and textures, which is crucial for the high-speed demands of the sport.

Overall, the primary challenges in single drone applications include limited battery life, which restricts operational duration, and the need for effective obstacle avoidance in varied environments. Additionally, limitations in data processing capabilities affect real-time decision-making and adaptability. Advanced technological solutions are essential to overcome these challenges, ensuring that single drones can operate efficiently and reliably in diverse scenarios and paving the way for future innovations.

\subsection{Multi-Drone Application}

\begin{figure}[!tb]
    \centering
    \begin{subfigure}{0.23\textwidth}
        \centering
        \includegraphics[width=0.93\textwidth, height=0.9\linewidth]{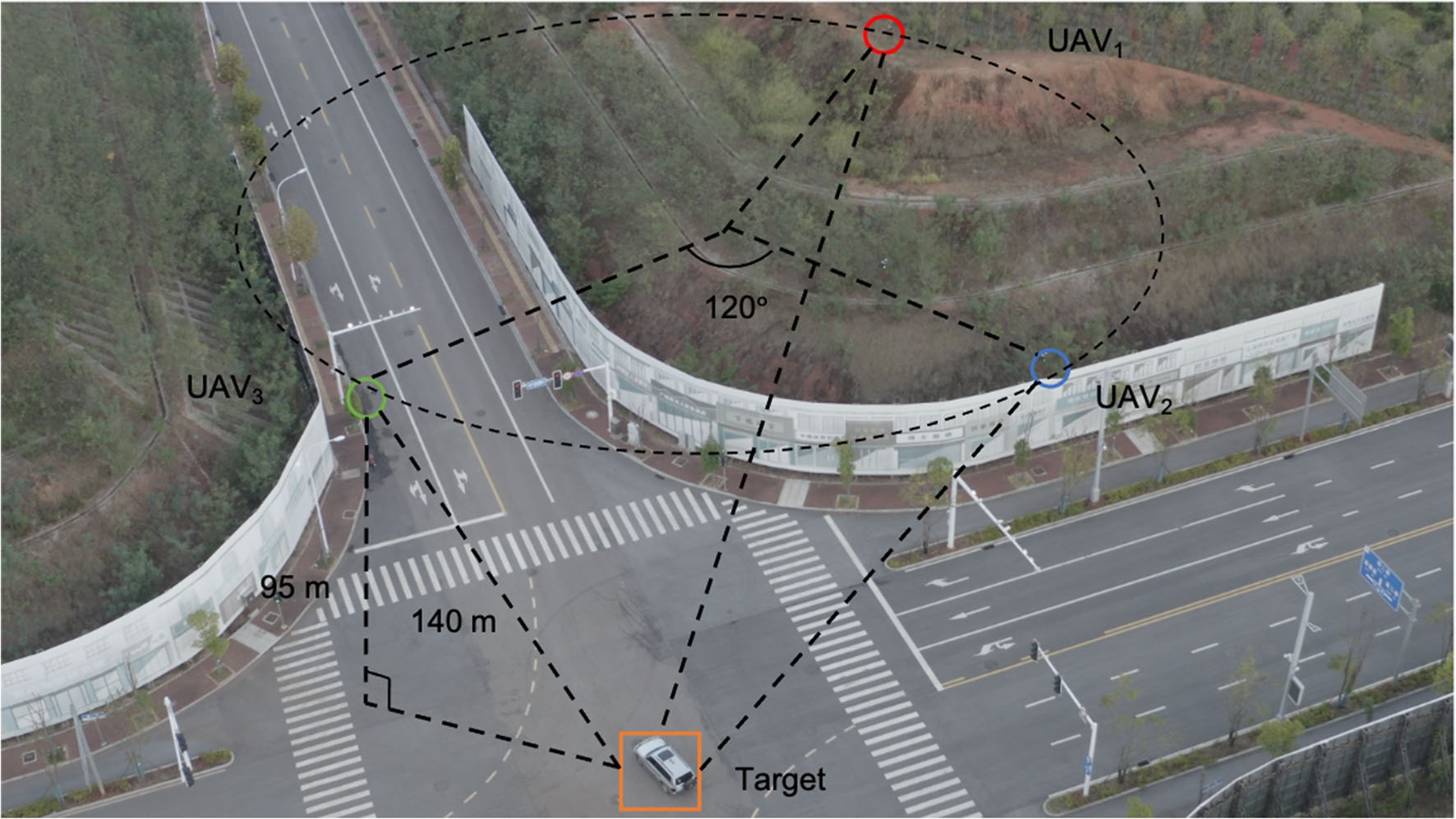}
        \caption{}
    \end{subfigure}%
    \begin{subfigure}{0.23\textwidth}
        \centering
        \includegraphics[width=0.93\textwidth, height=0.9\linewidth]{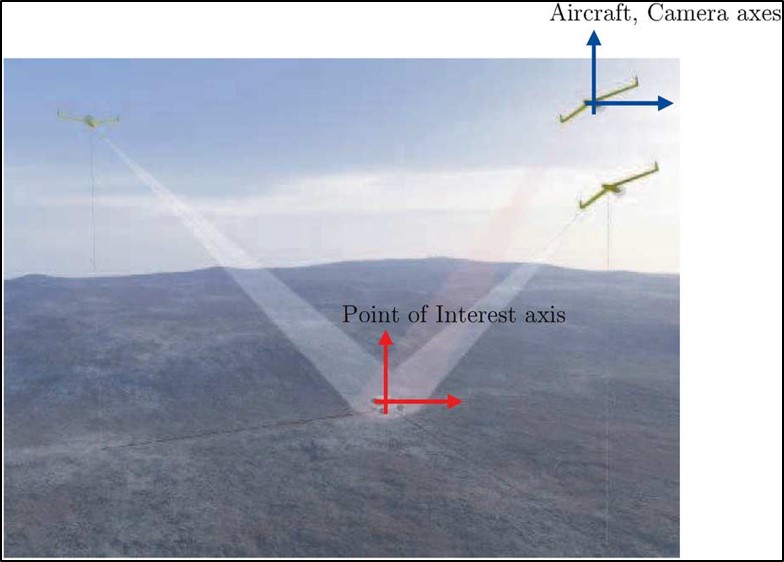}
        \caption{}
    \end{subfigure}
    \newline
    \begin{subfigure}{0.23\textwidth}
        \centering
        \includegraphics[width=0.93\textwidth, height=0.9\linewidth]{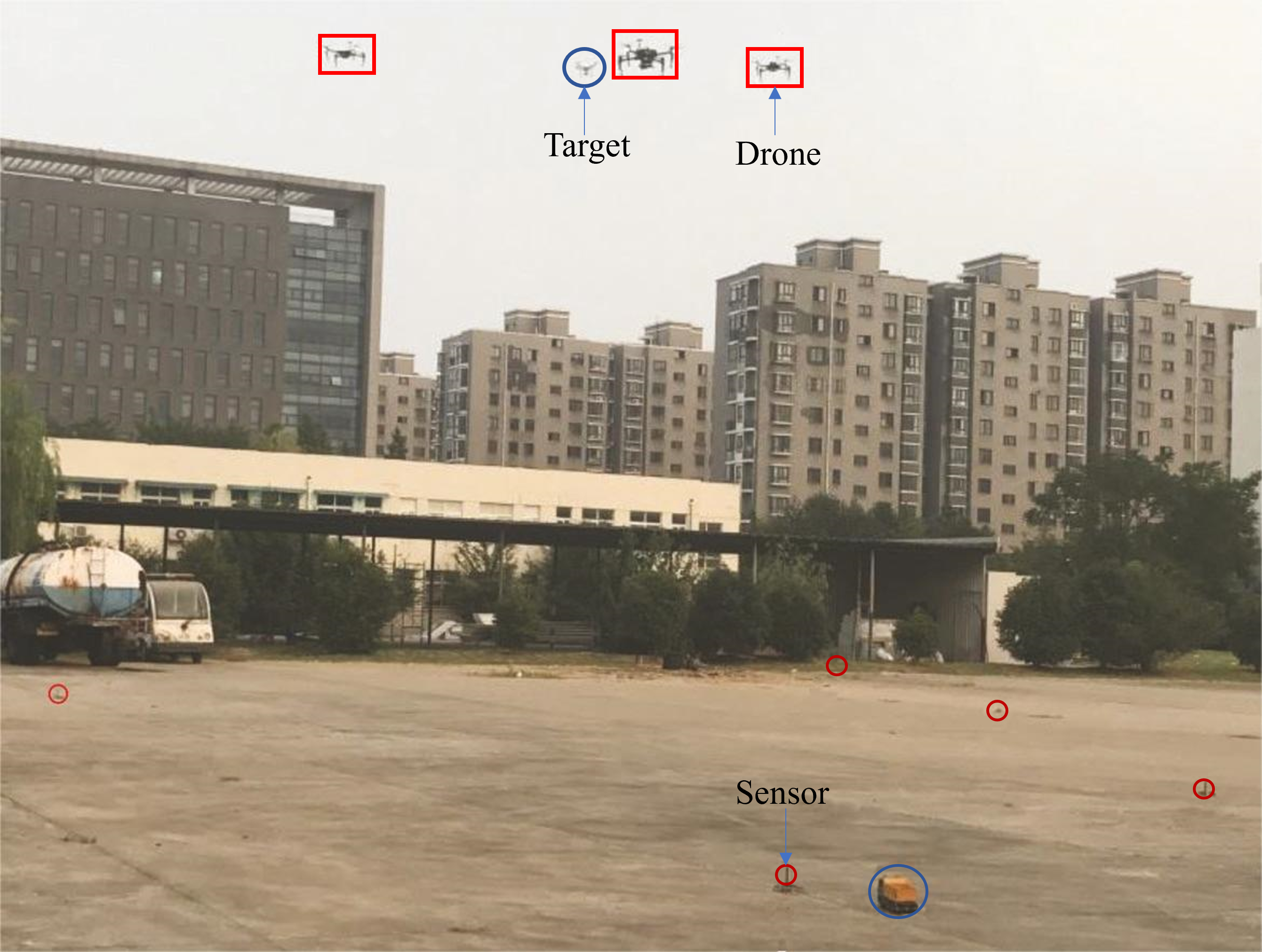}
        \caption{}
    \end{subfigure}%
    \begin{subfigure}{0.23\textwidth}
        \centering
        \includegraphics[width=0.93\textwidth, height=0.9\linewidth]{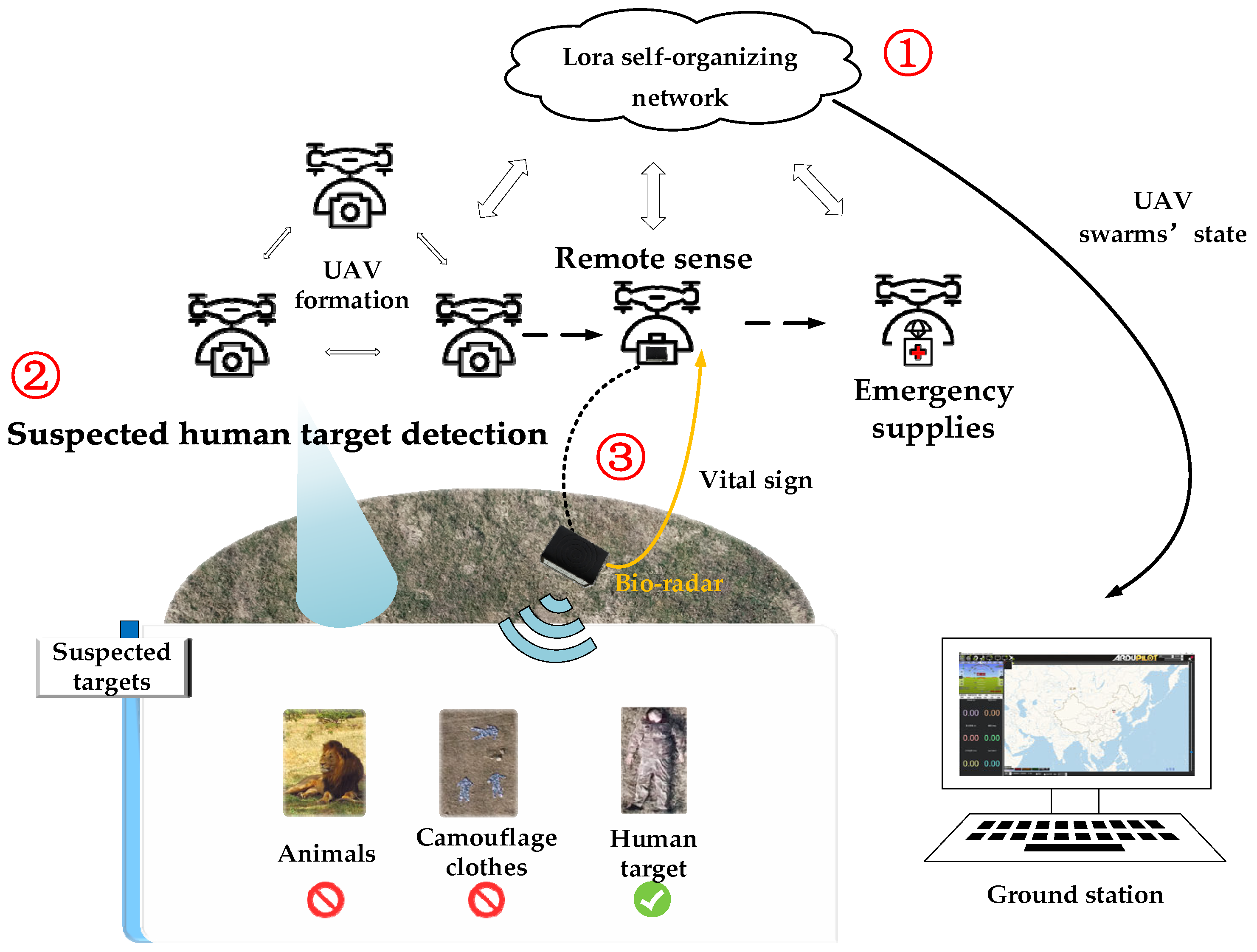}
        \caption{}
    \end{subfigure}
    \caption{Applications of multi-drone. (a) Coordinated Surveying \cite{lin2022end}; (b) Cooperative Tracking \cite{campbell2007cooperative}; (c) Synchronized Monitoring \cite{gu2018multiple}; (d) Disaster Response \cite{cao2022mission}.}
    \label{fig:multi_applications}
\end{figure}

% Multi-drone applications, such as coordinated surveying and disaster response, bring the added complexity of inter-drone communication and coordination. The main challenges here include maintaining stable and reliable communication links, collision avoidance between drones, and effective distribution of tasks to optimize the overall mission efficiency. Additionally, issues like signal interference and managing the flight paths of multiple drones simultaneously are significant hurdles.
While single drones offer convenience, their limited monitoring range has prompted interest in multi-drone collaboration. This approach seeks to overcome range limitations by leveraging the collective capabilities of multiple drones for broader, more efficient operations. Multi-drone applications, encompassing activities such as coordinated surveying \cite{loianno2015cooperative, tong2023multi, lin2022end, piasco2016collaborative}, cooperative tracking \cite{campbell2007cooperative, liu2022smart, farmani2017scalable}, synchronized monitoring \cite{gu2018multiple, jouhari2021distributed, yun2022cooperative}, and disaster response \cite{tang2018vision, scherer2015autonomous, xing2022multi, xiao2022collaborative, cao2022mission}, bring the added complexity of inter-drone communication, coordination and real-time data integration. These applications leverage the combined capabilities of multiple drones to achieve greater efficiency and coverage than single drone operations.

\subsubsection{Coordinated Surveying}In coordinated surveying, several challenges are prominent: merging diverse data from individual drones, and addressing computational demands in cooperative process. These challenges were tackled by some works. In \cite{loianno2015cooperative}, a monocular visual odometry algorithm was used to enable autonomous onboard control with cooperative localization and mapping. This work addressed the challenges of coordinating and merging different maps constructed by each drone platform and what’s more, the computational bottlenecks typically associated with 3D RGB-D cooperative SLAM. Micro-air vehicles also play an outstanding role in the field of coordinated surveying. Similarly, in \cite{piasco2016collaborative}, a sensor fusion scheme was proposed to improve the accuracy of localization in MAV fleets. Ensuring that each MAV effectively contributes to the overall perception of the environment poses a challenge addressed in this work. Both methods depend on effective collaboration and communication among multiple agents. However, maintaining stable communication between drones remains a critical and unresolved issue in multi-drone operations.

\subsubsection{Cooperative Tracking}In the field of multi-drone object tracking, navigating complex operational environments and overcoming limited communication bandwidth are significant challenges. These topics have also garnered significant research interest. In the study \cite{liu2022smart}, researchers developed a system for cooperative surveillance and tracking in urban settings, specifically tackling the issue of collision avoidance among drones. Additionally, another work by Farmani \textit{et al.} \cite{farmani2017scalable} explores a decentralized tracking system for drones, focusing on overcoming limited communication bandwidth and intermittent connectivity challenges. However, a persistent difficulty in this field is navigating complex external environments. Effective path planning and avoiding obstacles during multi-drone operations remain crucial challenges that require ongoing attention and innovation.

\subsubsection{Synchronized Monitoring}In synchronized monitoring missions with multiple drones, the focus is on how to effectively allocate tasks among drones and improve overall mission efficiency, as well as overcoming computational limitations. Gu \textit{et al.} in \cite{gu2018multiple} developed a small target detection and tracking model using data fusion, establishing a cooperative network for synchronized monitoring. This study also addresses task distribution challenges and efficiency optimization. Moreover, \cite{jouhari2021distributed} explores implementing deep neural network (DNN) models on computationally limited drones, focusing on reducing classification latency in collaborative drone systems. However, issues like collision avoidance in complex environments and signal interference in multi-drone systems are not comprehensively addressed in these studies.

\subsubsection{Disaster Response}In the domain of multi-drone systems for disaster response, challenges such as autonomous navigation, communication limitations, and real-time decision-making are paramount. To address these problems, Tang \textit{et al.} \cite{tang2018vision} introduced a tracking-learning-detection framework with an advanced flocking strategy for exploration and search missions. However, the study does not examine the crucial aspect of task distribution and optimization for enhancing mission efficiency in multi-drone disaster response. This shortcoming points to an essential area for further research, as effective task management is key to utilizing the full capabilities of multi-drone systems, particularly in the dynamic and urgent context of disaster scenarios.

The primary challenges in multi-drone applications include maintaining stable and reliable communication links, collision avoidance between drones, and effective distribution of tasks to optimize the overall mission efficiency. Additionally, issues like signal interference and managing the flight paths of multiple drones simultaneously are significant hurdles. Addressing these challenges is vital for unlocking the full potential of multi-drone systems, paving the way for advancements in various application domains.

\subsection{Heterogeneous Systems Application}
As the complexity and uncertainty of task scenarios escalate, it becomes increasingly challenging for a single robot or even a homogeneous multi-robot system to efficiently adapt to diverse environments. Consequently, in recent years, heterogeneous multi-robot systems (HMRS) have emerged as a focal point of research within the community \cite{rizk2019cooperative}. In the domain of UAV applications, HMRS mainly refers to communication-enabled networks that integrate various UAVs with other intelligent robotic platforms, such as unmanned ground vehicles (UGVs) and unmanned surface vehicles (USVs). This integration facilitates a diverse range of applications, leveraging the unique capabilities of each system to enhance overall operational efficiency. These systems execute a range of tasks, either individually or in a collaborative manner. The inherently heterogeneous scheduling approach of HMRS significantly enhances feasibility and adaptability, thereby effectively tackling a series of demanding tasks across different environments. Consequently, applications of HMRS are rapidly evolving; examples include but are not limited to localization and path planning\cite{niu2022unmanned, liu2022vision, li2016hybrid, zhang2022distributed}, precise landing \cite{niu2021vision, xu2020vision, hui2013autonomous}, and comprehensive inspection and detection \cite{kalinov2020warevision, minaeian2015vision, adaptable2021building}.

\begin{figure}[!tbp]
    \centering
    \begin{subfigure}{0.23\textwidth}
        \centering
        \includegraphics[width=0.93\textwidth, height=0.9\linewidth]{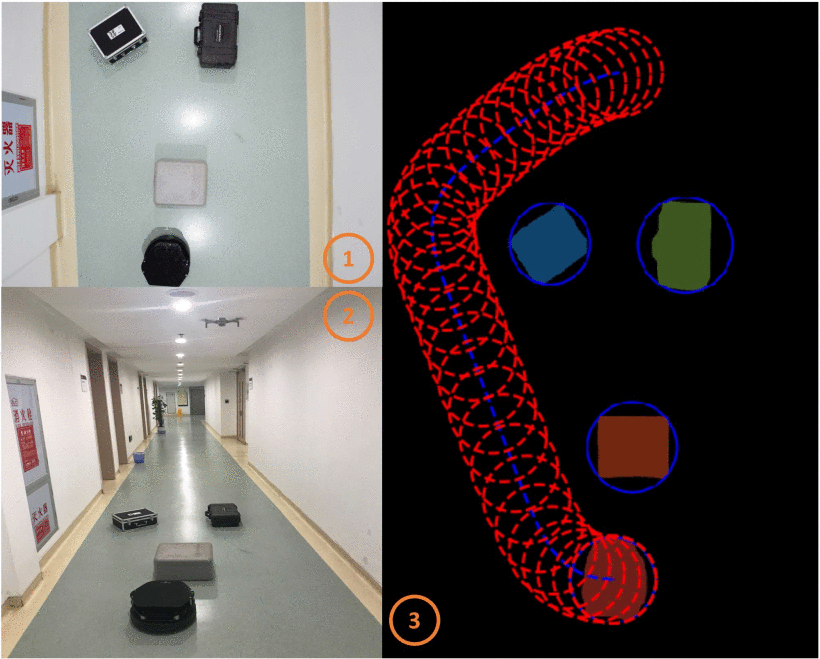}
        \caption{}
    \end{subfigure}%
    \begin{subfigure}{0.23\textwidth}
        \centering
        \includegraphics[width=0.93\textwidth, height=0.9\linewidth]{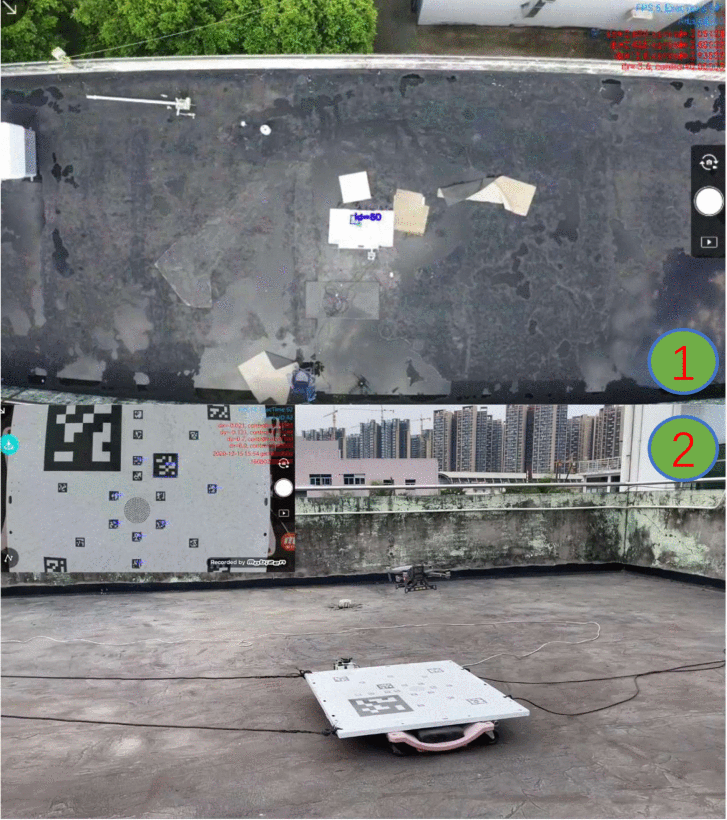}
        \caption{}
    \end{subfigure}
    \newline
    \begin{subfigure}{0.23\textwidth}
        \centering
        \includegraphics[width=0.93\textwidth, height=0.9\linewidth]{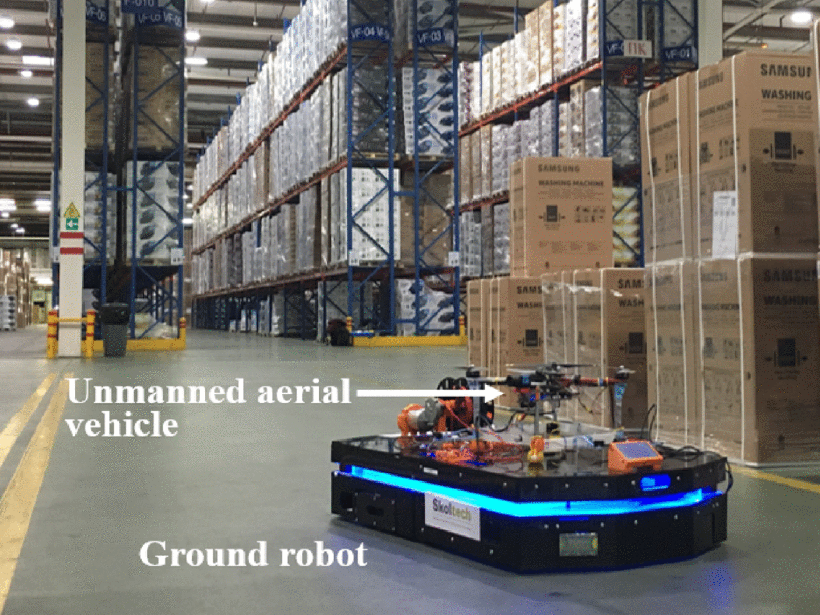}
        \caption{}
    \end{subfigure}%
    \begin{subfigure}{0.23\textwidth}
        \centering
        \includegraphics[width=0.93\textwidth, height=1.0\linewidth]{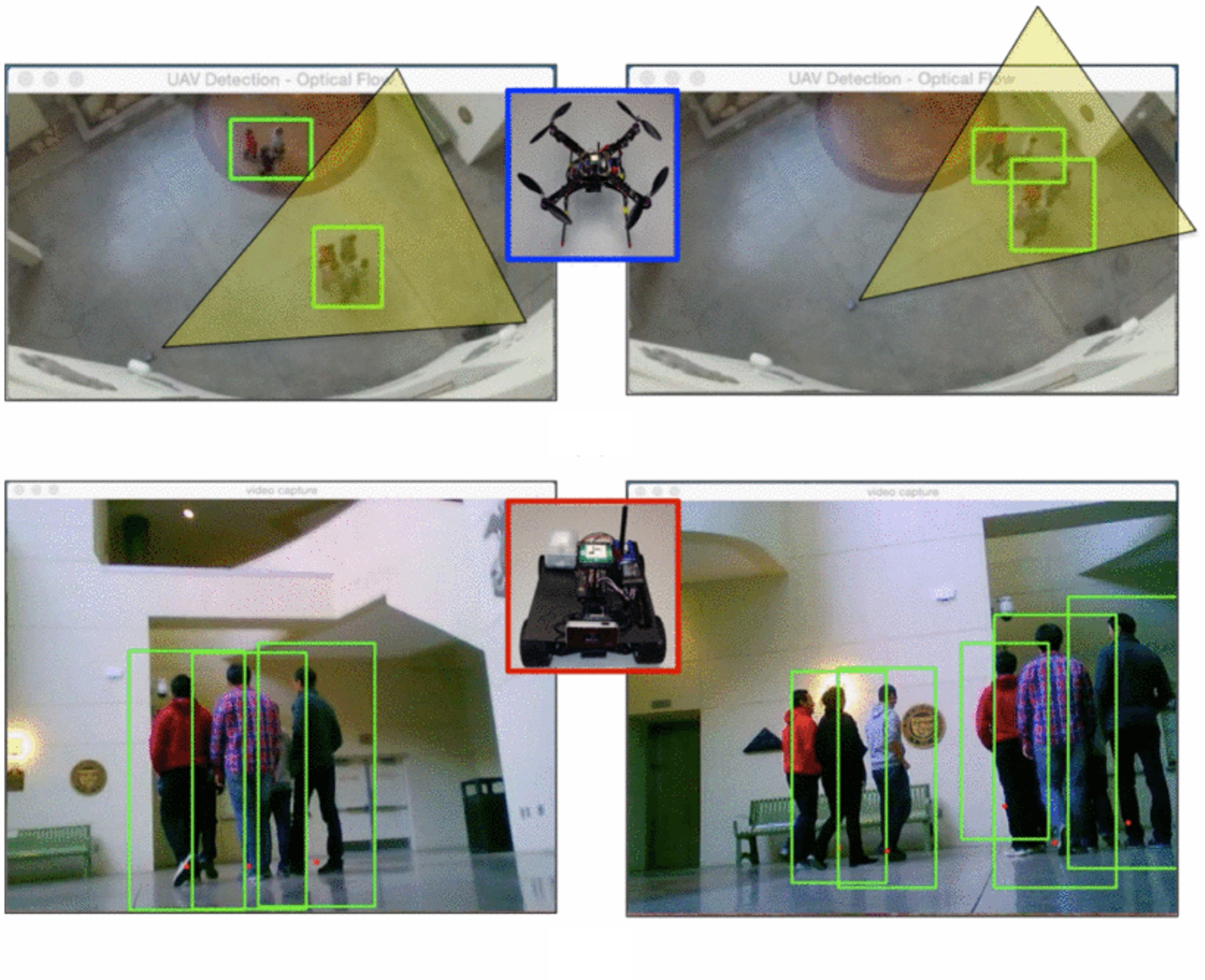}
        \caption{}
    \end{subfigure}
    \caption{Applications of heterogeneous systems. (a) UAV-UGV path planning \cite{niu2022unmanned}; (b) UAV-UGV precise landing \cite{niu2021vision}; (c) UAV-UGV inventory inspection \cite{kalinov2020warevision}; (d) UAV-UGV object detection \cite{minaeian2015vision}.}
    \label{HMRS}
\end{figure}

\subsubsection{Localization and Path Planning}
The UAV-UGV cooperation system can address the challenges of GPS denial and limited sensing range in UGVs. In this system, UAVs provide essential auxiliary information for UGV localization and path planning, relying solely on cost-effective cameras and wireless communication channels. Niu \cite{niu2022unmanned} introduced a framework wherein a single UAV served multiple UGVs, achieving optimal path planning based on aerial imagery. This approach surpassed traditional heuristic path planning algorithms in performance. Furthermore, Liu \textit{et al.} \cite{liu2022vision} presented a joint UAV-UGV architecture designed to overcome frequent target occlusion issues encountered by single-ground platforms. This architecture enabled accurate and dynamic target localization, leveraging visual inputs from UAVs.

\subsubsection{Precise Landing}
Given the potential necessity for battery recharging and emergency maintenance of UAVs during extended missions, the UAV-UGV heterogeneous system can facilitate UAV landings. This design reduces reliance on manual intervention while enhancing the UAVs' capacity for prolonged, uninterrupted operation. In the study \cite{niu2021vision}, a vision-based heterogeneous system was proposed to address the challenge of UAVs' temporary landings during long-range inspections. This system accomplished precise target geolocation and safe landings in the absence of GPS data by detecting QR codes mounted on UGVs. Additionally, Xu \textit{et al.} \cite{xu2020vision} illustrated the application of UAV heterogeneous systems for landing on USVs. A similar approach was explored for UAVs' target localization and landing, leveraging QR code recognition on USVs. Collectively, these studies underscored the feasibility and adaptability of the heterogeneous landing system across different platforms.

\subsubsection{Inspection and Detection}
Heterogeneous UAV systems present an effective solution to overcome the limitations of background clutter and incoherent target interference often encountered in single-ground vision detection platforms. By leveraging the expansive field of view and swift scanning capabilities of UAVs, in conjunction with the endurance and high accuracy of UGVs, such heterogeneous systems can achieve time-efficient and accurate target inspection and detection in specific applications. For instance, Kalinov \textit{et al.} \cite{kalinov2020warevision} introduced a heterogeneous inventory management system, pairing a ground robot with a UAV. In this system, the ground robot determined motion trajectories by deploying the SLAM algorithm, while the UAV, with its high maneuverability, was tasked with scanning barcodes. Furthermore, Pretto \cite{adaptable2021building} developed a heterogeneous farming system to enhance agricultural automation. This innovative system utilized the aerial perspective of the UAV to assist in farmland segmentation and the classification of crops from weeds, significantly contributing to the advancement of automated farming practices.

To sum up, most of the applications above primarily focus on single UAV to single UGV or single UAV to multiple UGV configurations, with few scenarios designed for multiple UAVs interacting with multiple UGVs. It is evident that there remains significant research potential in the realm of vision-based, multi-agent-to-multi-agent heterogeneous systems. Key areas such as communication and data integration within heterogeneous systems, coordination and control in dynamic and unpredictable environments, and individual agents' autonomy and decision-making capabilities warrant further exploration.

\section{Open Questions and Potential Solutions} \label{sec: question}
Despite significant advancements in the domain of vision-based learning for drones, numerous challenges remain that impede the pace of development and real-world applicability of these methods. These challenges span various aspects, from data collection and simulation accuracy to operational efficiency and safety concerns.

\subsection{Dataset}
A major impediment in the field is the absence of a comprehensive, public dataset analogous to Open X-Embodiment \cite{open_x_embodiment_rt_x_2023} in robotic manipulation. This unified dataset should ideally encompass a wide range of scenarios and tasks to facilitate generalizable learning. The current reliance on domain-specific datasets like “Anti-UAV” \cite{9615243} and “SUAV-DATA” \cite{zhao2023tgc} limits the scope and applicability of research. A potential solution is the collaborative development of a diverse, multi-purpose dataset by academic and industry stakeholders, incorporating various environmental, weather, and lighting conditions.

\subsection{Simulator}
While simulators are vital for training and validating vision-based drone models, their realism and accuracy often fall short of replicating real-world complexities. This gap hampers the transition from simulation to actual deployment. Meanwhile, there is no unified simulator covering most of the drone tasks, resulting in repetitive domain-specific simulator development\cite{shah2018airsim, song2021flightmare, panerati2021learning}. Drawing inspiration from the self-driving car domain, the integration of off-the-shelf and highly flexible simulators such as CARLA\cite{dosovitskiy2017carla} could be a solution. These simulators, known for their advanced features in realistic traffic simulation and diverse environmental conditions, can provide more authentic and varied data for training. Adapting such simulators to drone-specific scenarios could greatly enhance the quality of training and testing environments.

\subsection{Sample Efficiency}
Enhancing sample efficiency in machine learning models for drones is crucial, particularly in environments where data collection is hazardous or impractical. Even though simulators are available for generating training data, there are still challenges in ensuring the realism and diversity of these simulated environments. The gap between simulated and real-world data can lead to performance discrepancies when models are deployed in actual scenarios. Developing algorithms that leverage transfer learning\cite{zhuang2020comprehensive}, few-shot learning\cite{wang2020generalizing}, and synthetic data generation \cite{fonseca2023tabular} could provide significant strides in learning efficiently from limited datasets. These approaches aim to bridge the gap between simulation and reality, enhancing the applicability and robustness of machine learning models in diverse and dynamic real-world situations.

\subsection{Inference Speed}
Balancing inference speed with accuracy is a critical challenge for drones operating in dynamic environments. The key lies in optimizing machine learning models for edge computing, enabling drones to process data and make decisions swiftly. Techniques like model pruning \cite{liu2018rethinking, jiang2022model}, quantization \cite{zhou2018adaptive, deng2020model}, distillation \cite{wang2021knowledge} and the development of specialized hardware accelerators can play a pivotal role in this regard.

\subsection{Real World Deployment}
Transitioning from controlled simulation environments to real-world deployment (Sim2Real) involves addressing unpredictability in environmental conditions, regulatory compliance, and adaptability to diverse operational contexts. Domain randomization \cite{tobin2017domain} tries to address the Sim2Real issue in a certain way but is limited to predicted scenarios with known domain distributions. Developing robust and adaptive algorithms capable of on-the-fly learning and decision-making, along with rigorous field testing under varied conditions, can aid in overcoming these challenges.

\subsection{Embodied Intelligence in Open World}
Existing vision-based learning methods for drones require explicit task descriptions and formal constraints, while in an open world, it is hard to provide all necessary formulations at the beginning to find the optimal solution. For instance, in a complex search and rescue mission, the drone can only find the targets first and conduct rescue based on the information collected. In each stage, the task may change, and there is no prior explicit problem at the start. Human interactions are necessary during this mission. With large language models and embodied intelligence, the potential of drone autonomy can be greatly increased. Through interactions in the open world\cite{gupta2021embodied, gong2023mindagent} or provide few-shot imitation \cite{bhoopchand2023learning}, vision-based learning can emerge with full autonomy for drone applications.

\subsection{Safety and Security}
Ensuring the safety and security of drone operations is paramount, especially in densely populated or sensitive areas. This includes not only physical safety but also cybersecurity concerns\cite{xiao2022cyber,9596578}. The security aspect extends beyond data protection, including the resilience of drones to adversarial attacks. Such attacks could take various forms, from signal jamming to deceptive inputs aimed at misleading vision-based systems and DRL algorithms \cite{ilahi2021challenges}. Addressing these concerns requires a multifaceted approach. Firstly, incorporating advanced cryptographic techniques ensures data integrity and secure communication. Secondly, implementing anomaly detection systems can help identify and mitigate unusual patterns indicative of adversarial interference. Moreover, improving the robustness of learning models against adversarial attacks and investigating the explainability of designed models \cite{rawal2021recent, vouros2022explainable} are imperative. Lastly, regular updates and patches to the drone's software, based on the latest threat intelligence, can fortify its defenses against evolving cyber threats.

\section{Conclusion} \label{sec:conclusion}
This comprehensive survey has thoroughly explored the rapidly growing field of vision-based learning for drones, particularly emphasizing their evolving role in multi-drone systems and complex environments such as search and rescue missions and adversarial settings. The investigation revealed that drones are increasingly becoming sophisticated, autonomous systems capable of intricate tasks, largely driven by advancements in AI, machine learning, and sensor technology. The exploration of micro and nano drones, innovative structural designs, and enhanced autonomy stand out as key trends shaping the future of drone technology. Crucially, the integration of visual perception with machine learning algorithms, including deep reinforcement learning, opens up new avenues for drones to operate with greater efficiency and intelligence. These capabilities are particularly pertinent in the context of object detection and decision-making processes, vital for complex drone operations. The survey categorized vision-based control methods into indirect, semi-direct, and end-to-end methods, offering a nuanced understanding of how drones perceive and interact with their environment. Applications of vision-based learning drones, spanning from single-agent to multi-agent and heterogeneous systems, demonstrate their versatility and potential in various sectors, including agriculture, industrial inspection, and emergency response. However, this expansion also brings forth challenges such as data processing limitations, real-time decision-making, and ensuring robustness in diverse operational scenarios.

The survey highlighted open questions and potential solutions in the field, stressing the need for comprehensive datasets, realistic simulators, improved sample efficiency, and faster inference speeds. Addressing these challenges is crucial for the effective deployment of drones in real-world scenarios. Safety and security, especially in the context of adversarial environments, remain paramount concerns that need ongoing attention. While significant progress has been made in vision-based learning for drones, the journey towards fully autonomous, intelligent, and reliable systems, even AGI in the physical world, is ongoing. Future research and development in this field hold the promise of revolutionizing various industries, pushing the boundaries of what's possible with drone technology in complex and dynamic environments.

\bibliographystyle{IEEEtran}
\bibliography{IEEEabrv,xiao}

\end{document}